\documentclass{article} 
\usepackage{iclr2027_conference,times}

\usepackage{amsmath,amsfonts,bm}

\def\eqref#1{equation~\ref{#1}}

\def\1{\bm{1}}

\def\vh{{\bm{h}}}

\def\vm{{\bm{m}}}

\def\vx{{\bm{x}}}

\DeclareMathAlphabet{\mathsfit}{\encodingdefault}{\sfdefault}{m}{sl}
\SetMathAlphabet{\mathsfit}{bold}{\encodingdefault}{\sfdefault}{bx}{n}

\usepackage{url}

\usepackage{listings}
\usepackage{xcolor}

\definecolor{codegreen}{rgb}{0,0.6,0}
\definecolor{codegray}{rgb}{0.5,0.5,0.5}
\definecolor{codepurple}{rgb}{0.58,0,0.82}
\definecolor{backcolour}{rgb}{0.96,0.96,0.96}
\definecolor{prompttext}{rgb}{0.2,0.2,0.2}

\newcounter{algocount}
\newcounter{promptcount}

\lstdefinestyle{pythonstyle}{
    language=Python,
    backgroundcolor=\color{backcolour},   
    commentstyle=\color{codegreen},
    keywordstyle=\color{magenta},
    numberstyle=\tiny\color{codegray},
    stringstyle=\color{codepurple},
    basicstyle=\ttfamily\footnotesize,
    breakatwhitespace=false,         
    breaklines=true,                 
    captionpos=t,                    
    keepspaces=true,                 
    numbers=left,                    
    numbersep=5pt,                  
    showspaces=false,                
    showstringspaces=false,
    tabsize=2,
    frame=single
}

\lstdefinelanguage{llmprompt}{
    keywords={User:, Assistant:, System:, Input:, Output:},
    keywordstyle=\bfseries\color{blue!60!black},
    sensitive=true
}
\lstdefinestyle{promptstyle}{
    language=llmprompt,
    basicstyle=\ttfamily\footnotesize\color{prompttext},
    backgroundcolor=\color{white},
    breaklines=true,
    frame=single,
    captionpos=t,
    numbers=left,
    numberstyle=\tiny\color{codegray},
    columns=fullflexible
}

\lstnewenvironment{pyalgorithm}[2][]{%
    \setcounter{lstlisting}{\value{algocount}}%
    \lstset{style=pythonstyle, caption={#2}, #1}%
}{%
    \stepcounter{algocount}%
}

\lstnewenvironment{llmprompt}[2][]{%
    \setcounter{lstlisting}{\value{promptcount}}%
    \lstset{style=promptstyle, caption={#2}, #1}%
}{%
    \stepcounter{promptcount}%
}

\usepackage{microtype}
\usepackage{graphicx}
\usepackage{subcaption}
\usepackage{booktabs} 
\usepackage{hyperref}
\usepackage{amsmath}
\usepackage{amssymb}
\usepackage{mathtools}
\usepackage{amsthm}
\usepackage{tcolorbox}
\usepackage{algorithm}
\usepackage{algpseudocode}

\newcommand{\name}[1]{{TRACE}}

\title{TRACE: Tackling Real-World Resource \\ Assignment Problems via Agentic \\ Heuristic Design}

\author{
Jose A. Ayala-Romero\textsuperscript{1}, 
Andres Garcia-Saavedra\textsuperscript{1}, 
Xavier Costa-Perez\textsuperscript{2,1} \\
\textsuperscript{1}NEC Laboratories Europe \\
\textsuperscript{2}i2CAT Foundation and ICREA \\
}

\iclrfinalcopy 

\begin{document}

\maketitle

\begin{abstract}
Dynamic resource assignment, the real-time allocation of task streams to heterogeneous processing nodes, is the backbone of modern computing infrastructure. While learning-based schedulers excel in research, industrial deployments still rely on hand-written rules that operators can read, audit, and execute within tight latency budgets. LLM-based Automatic Heuristic Design (AHD) promises to automate writing such rules. However, existing AHD frameworks were developed for combinatorial problems fully specified to the LLM, and they learn only from a scalar fitness score. In real systems, the behaviour that determines a good heuristic, such as processor speeds or power consumption, is unknown a priori: the score reveals \textit{which} heuristic performs better, but not \textit{why}. This missing information is recorded in the system logs that every evaluation produces.
Exploiting it is non-trivial: logs are massive and noisy, the relevant signals depend on the objective, and their content and format vary across hardware and software stacks, so they can neither be fed to an LLM as is nor processed by a fixed parser.
We propose \name{}, which couples an evolutionary AHD loop with an agentic knowledge-extraction workflow. A Reasoner agent analyzes the log schema in light of the objective and formulates hypotheses about the system dynamics; a Coder agent writes and executes schema-specific code to test them, producing insights or executable tools for the evolved heuristics.
We evaluate \name{} on a synthetic cloud benchmark and a 5G vRAN scenario built from industrial testbed measurements and operational traffic traces. \name{} consistently outperforms state-of-the-art AHD methods in resource assignment problems and yields more auditable heuristics at under 2\% overhead.
\end{abstract}

\section{Introduction}

Dynamic task scheduling is the operational backbone of modern industrial infrastructure: incoming streams of tasks must be allocated in real time to a limited set of processing nodes. In cloud computing data centers, orchestrators assign heterogeneous workloads (e.g., AI training jobs or virtual machines) across thousands of servers \citep{cloud_survey_1, cloud_suvery_2}. In wired and wireless networks, routers and base stations act as computing nodes that continuously process streams of data packets \citep{cloudric, aslan2024fair}. Inefficient scheduling leads to increased latency, energy waste, and service violations.

Despite decades of research, production schedulers remain predominantly hand-written rules. Classical formulations such as the Dynamic Job-Shop Scheduling Problem (DJSSP) \citep{djssp_1, djssp_2} assume known processing times and unbounded queues, which rarely match deployed systems. Deep reinforcement learning (DRL) relaxes these assumptions \citep{resource_microsoft, task_assignment}, but its policies are hard for operators to inspect, certify, and modify, and can degrade under the simulation-to-deployment shift. Operators therefore favor rules they can read, audit, and execute within tight latency budgets \citep{agora}. Writing \textit{good} rules, however, requires expert knowledge of the specific hardware, which must be re-acquired for every new deployment.

LLM-based Automatic Heuristic Design (AHD) can automate the writing of such rules. Frameworks such as FunSearch \citep{funsearch}, EoH \citep{eoh}, ReEvo \citep{reevo}, HSEvo \citep{hsevo}, and AlphaEvolve \citep{novikov2025alphaevolve}, which has even improved a production data-center scheduling heuristic, evolve programs in a loop: an LLM proposes code, an evaluator returns a scalar fitness, and the best candidates seed the next generation. These approaches, however, share a blind spot: \textit{the only feedback is a scalar}. This is largely sufficient on the combinatorial benchmarks where AHD has been developed (e.g., TSP, flow-shop, or online bin packing), because the LLM is given the full problem specification. In real resource-assignment systems, instead, the quantities that determine which heuristic is good, such as per-node processing times, energy profiles, and queue capacities, are \textit{unknown to the designer a priori}. Scalar-feedback search can learn \textit{which} heuristic scores better, but not \textit{why}.

\begin{figure}[t!]
\centering
    \includegraphics[width=\linewidth]{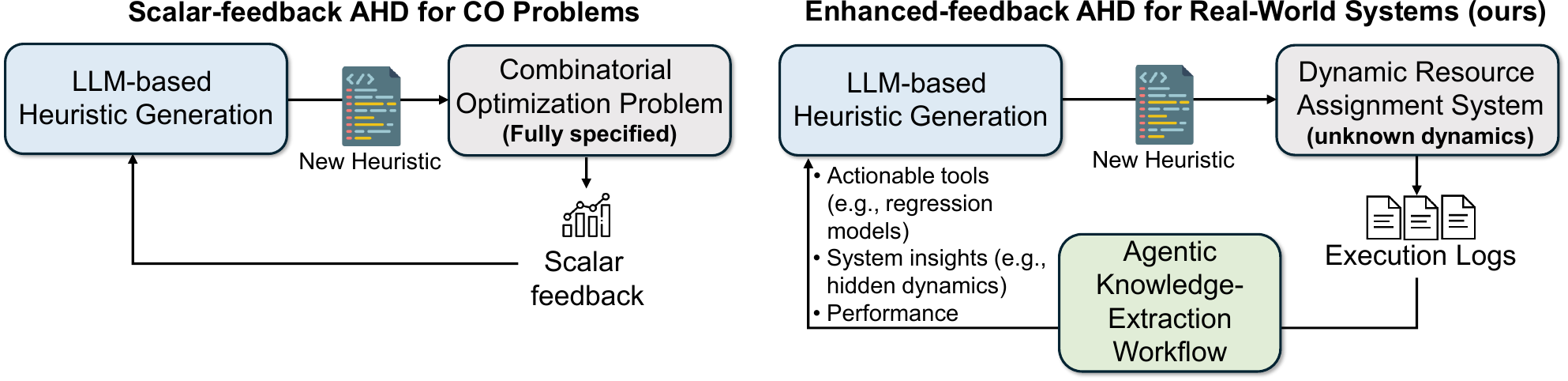}
    \caption{Scalar-feedback AHD (left) learns \textit{what} works from a single fitness value. \name{} (right) distills system logs into knowledge about \textit{why} it works (e.g., per-processor speed and energy profiles) and feeds it back into heuristic generation.}
    \label{fig:scope}
    \vspace{-3mm}
\end{figure}

Consider a cluster in which one node is much faster than the rest but also far more power-hungry. An energy-efficient policy should use this accelerator for very heavy tasks or when the system is saturated, which requires knowing each processor's performance and energy profile. Scalar-feedback search can only stumble upon this behavior by trial and error, yet the needed information is recorded in the system logs during every evaluation. \textbf{Our key insight is that these logs can turn heuristic search from blind trial and error into informed, data-driven design} (Fig.~\ref{fig:scope}).

Exploiting logs is nevertheless non-trivial. ($i$)~Raw logs can reach gigabytes and are dominated by irrelevant entries, so they cannot simply be placed in an LLM prompt. ($ii$)~What is relevant depends on the objective: minimizing latency or energy calls for different metrics, and which ones matter is often unclear in advance. ($iii$)~Logs are heterogeneous in content and format: the available metrics depend on the hardware, and field names, units, and structure vary across systems, so a fixed, hand-written parser does not transfer. Moreover, reflection-based AHD methods (e.g., ReEvo) reason only over the fitness history, not over the system behavior behind it. Extracting useful knowledge thus requires reasoning jointly about the objective and the log schema, and writing code tailored to both.
To this end, we propose \name{}, which augments LLM-based AHD with an \textit{agentic knowledge-extraction workflow}: a \textit{Reasoner} agent identifies which log information is relevant to the objective, and a \textit{Coder} agent writes and executes code to extract it, producing either an insight into system behavior or a tool. This knowledge is fed back into the heuristic-generation loop to inform each new candidate.

We evaluate \name{} on ($i$)~a synthetic cloud-computing benchmark with up to $50$ heterogeneous servers that tests scalability, and ($ii$)~an energy-efficient 5G virtualized RAN scenario driven by measurements from an industrial-grade testbed and traffic traces from an operational network. In the cloud benchmark, \name{} is the only method that achieves zero task loss with $50$ servers, reducing cost by more than an order of magnitude relative to the best AHD baseline. In the 5G scenario, it matches the carrier-grade reliability of the industry-standard policy (0\% loss) while consuming 65\% and 25\% less energy than the industry standard and SoTA AHD approaches, respectively. 
We evaluate the auditability of \name{}'s heuristics using an LLM-as-a-judge approach, where they obtained the heuristics highest in semantic grounding and auditability, as their logic is expressed in terms of physical quantities (predicted processing times, per-processor thresholds) rather than the opaque arithmetic the baselines converge to. All these gains come at a 1.9\% overhead relative to EoH at an equal generation budget.

\section{Problem Formulation}\label{sec:prob_form}

\subsection{Dynamic Resource Assignment}
We consider a system with $N$ heterogeneous computing nodes $\mathcal{N}=\{1,\dots,N\}$ that receives a stream of tasks. Each task $i$ arrives at time $\tau_i$ and is described by a feature vector $\vx_i \in \mathbb{R}^d$ (e.g., token count, or inferred CPU and memory requirements). We make no assumptions on the arrival process or on the distribution of $\vx_i$.
If task $i$ is assigned to node $n$, it waits in the node's FIFO queue for $t^{\text{wait}}_{i,n}$ and is then processed for $t_{i,n} \sim T^{(n)}(\cdot \mid \vx_i)$, incurring a resource cost $r_{i,n} \sim R^{(n)}(\cdot \mid \vx_i)$ (e.g., energy). Node $n$ has queue capacity $Q^{(n)}$: a task assigned to a full queue is dropped. Systems may impose additional drop conditions, such as a deadline $D$ on $t^{\text{wait}}_{i,n}+t_{i,n}$ (Sec.~\ref{sec:5g}).
Critically, the functions defining the system behavior $\{T^{(n)}, R^{(n)}, Q^{(n)}\}_{n=1}^N$ are are noisy, arbitraty and \textit{unknown a priori}.
In industrial settings, they depend on undocumented hardware characteristics, firmware, and power-management settings.
A \textit{heuristic} is a program $g: \mathbb{R}^d \to \mathcal{N}$ that assigns each task to node $n_i = g(\vx_i)$. The per-task cost is
\begin{equation}\label{eq:cost}
C_i(g) =
\begin{cases}
\mathcal{P}_{i,n_i} & \text{if task } i \text{ is processed,} \\
\Gamma & \text{if task } i \text{ is dropped,}
\end{cases}
\end{equation}
where $\mathcal{P}_{i,n_i}$ is an application-specific performance term and $\Gamma \gg \mathcal{P}$ makes reliability the primary objective. Because waiting times and drops depend on earlier assignments, $C_i$ depends on $g$ through the whole trajectory. Given a set of workloads $\mathcal{W}$, the quality of $g$ is
$J(g) = \mathbb{E}_{W \sim \mathcal{W}}\big[\tfrac{1}{|W|}\sum_{i\in W} C_i(g)\big]$.
We instantiate $\mathcal{P}$ as latency, $\mathcal{P}_{i,n}=t^{\text{wait}}_{i,n}+t_{i,n}$ (cloud, Sec.~\ref{sec:cloud_computing}), and as energy, $\mathcal{P}_{i,n}=r_{i,n}$ with a deadline $D$ (5G vRAN, Sec.~\ref{sec:5g}).

\subsection{Problem formulation: Heuristic design from execution feedback}
We seek $g^\star = \arg\min_{g\in\mathcal{G}} J(g)$ over a space $\mathcal{G}$ of programs, without access to $\{T^{(n)},R^{(n)},Q^{(n)}\}$. Instead, the designer can \textit{evaluate} any candidate $g$ on the system or a simulator. Each evaluation returns the scalar fitness $J(g)$ and an execution log $\Lambda(g)=\{(\tau_i,\vx_i,n_i,\vm_i)\}_{i}$, where $\vm_i$ contains the metrics that system $S$ records for task $i$ (e.g., processing time, drop status, queue occupancy, energy).
Existing AHD methods search $\mathcal{G}$ using only $J(g)$. We study how to also exploit $\Lambda(g)$. Three challenges make this non-trivial:
($a$) $|\Lambda(g)|$ grows with the number of tasks and evaluations, so logs cannot be placed in an LLM context;
($b$) which components of $\vm_i$ are informative about $J$ depends on the instantiation of $\mathcal{P}$, and even informative components are raw per-task records that must be distilled (e.g., into per-node statistics or predictive models) before a heuristic can use them;
($c$) the recorded metrics and log format depend on the system, so extraction cannot be fixed in advance.
Distilled knowledge may be a textual insight or an executable tool (e.g., a predictor of $t_{i,n}$); we allow heuristics to call such tools $\mathcal{K}$, i.e., $g \in \mathcal{G}(\mathcal{K})$.

\section{Methodology}\label{sec:method}

\begin{figure*}[t!]
\centering
    \includegraphics[width=0.99\textwidth]{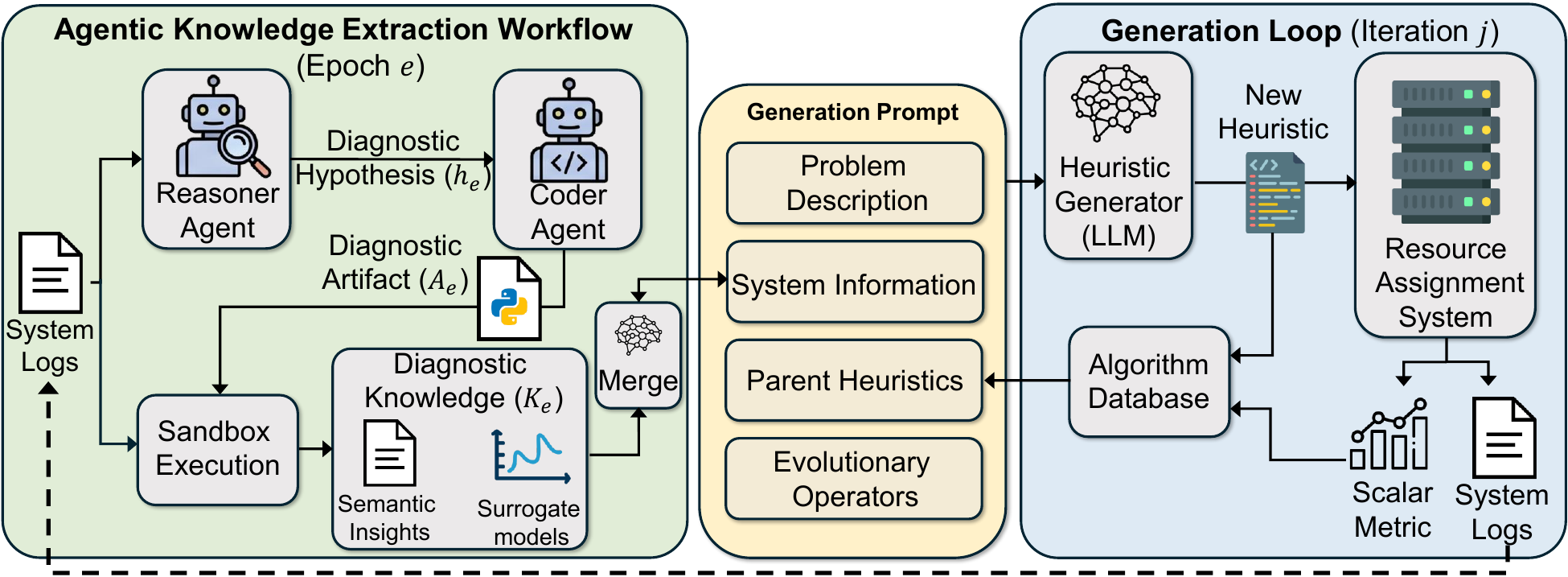}
    \caption{Overview of \name{}. \textbf{Right:} the generation loop evolves heuristics; each evaluation returns a fitness $J(g)$, stored in the algorithm database, and an execution log $\Lambda(g)$, stored in a log buffer. \textbf{Left:} at the start of each epoch $e$, the Agentic Workflow distills the log buffer into Diagnostic Knowledge $K_e$, merged into the System Information of the generation prompt.}
    \label{fig:methodology}
    \vspace{-4mm}
\end{figure*}

\name{} exploits both the fitness $J(g)$ and the execution logs $\Lambda(g)$ of every evaluation. It couples two nested loops (Fig.~\ref{fig:methodology}; Algorithm~\ref{alg:trace}): an evolutionary \textit{generation loop} that proposes and evaluates heuristics, and an \textit{Agentic Knowledge Extraction Workflow} that runs at the start of each epoch and distills logs into textual insights and executable tools $\mathcal{K}$, exposed to the heuristics through the generation prompt. The design targets the three challenges of Sec.~\ref{sec:prob_form}: ($a$)~the LLM agents only see a short log excerpt or the log schema, while the full log is processed by generated code; ($b$)~hypotheses are formulated in light of the optimization objective and specify how the relevant metrics are distilled into insights or tools; and ($c$)~the extraction code is written for the log schema of each system.

\subsection{Heuristic generation loop}\label{sec:gen_loop}

The evolutionary backbone of \name{} follows the principles described by \citet{eoh}. The algorithm database $\mathcal{D}$ holds a fixed-size population of heuristics and their fitness. The search runs over epochs $e=1,\dots,E$, each comprising $I$ generation steps $j=1,\dots,I$. At each step, the LLM receives a \textit{generation prompt} (Appendix~\ref{app:prompts}) with four parts: ($i$)~a problem description $\Pi$, including the objective and the heuristic interface; ($ii$)~the \textit{System Information} $\sigma_e$, a textual description of what is known about the system; ($iii$)~$m=2$ parent heuristics sampled from $\mathcal{D}$ with a bias toward high fitness; and ($iv$)~one evolutionary operator (e.g., crossover or mutation). The generated heuristic $g$ is evaluated, returning its fitness $J(g)$ (\eqref{eq:cost}) and log $\Lambda(g)$. Invalid programs (e.g., wrong signature or runtime errors) are discarded. Valid ones are inserted into $\mathcal{D}$, and their log records are appended to a bounded \textit{log buffer} $\Lambda$ that discards the oldest records first. Accumulating records across evaluations avoids biasing the extracted knowledge toward any single heuristic. The buffer is initialized with the log of a random policy, so that all nodes are observed from the start.

The System Information starts from a minimal description $\sigma_0$ (e.g., ``the relationship between task features and processing time is unknown'') and is enriched by the Agentic Workflow. When the workflow produces a tool, $\sigma_e$ documents its signature and usage, and the tool is made available at evaluation time, so heuristics can call it, i.e., $g\in\mathcal{G}(\mathcal{K})$. Without the Agentic Workflow, $\sigma_e=\sigma_0$ and \name{} reduces to EoH, which isolates the contribution of the logs (Sec.~\ref{sec:eval1}).

\subsection{Agentic Workflow}\label{sec:agentic_workflow}

\begin{figure*}[t!]
\centering
    \includegraphics[width=0.95\textwidth]{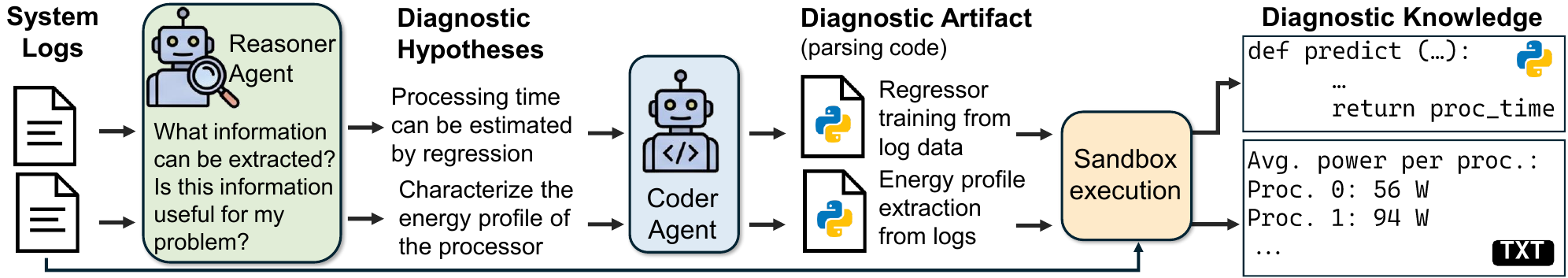}
    \caption{Two runs of the Agentic Workflow. \textbf{Top:} the Reasoner hypothesizes that processing time can be predicted from task features; the Coder writes code that trains a regressor on the logs, yielding a callable surrogate \texttt{predict()}. \textbf{Bottom:} characterizing the energy profile of each processor yields a semantic insight (per-processor average power).}
    \label{fig:agentic_workflow}
    \vspace{-4mm}
\end{figure*}

The Agentic Workflow converts the log buffer into knowledge in three steps (Fig.~\ref{fig:agentic_workflow}) that involve three objects:
a \textbf{Diagnostic Hypothesis} $h_e$, a short natural-language statement of what could be learned from the logs;
a \textbf{Diagnostic Artifact} $A_e$, a program that realizes $h_e$ on the logs; and
the \textbf{Diagnostic Knowledge} $K_e$ obtained by executing $A_e$.
$K_e$ is either a \textit{Semantic Insight}, i.e., a textual description of a behavioural pattern (e.g., average power per processor), or a \textit{Surrogate Model}, i.e., a callable estimator of an unknown system quantity.

\textbf{Reasoner.} The Reasoner agent formulates $h_e$ from the problem description $\Pi$ (and thus the objective, challenge~$b$), the current System Information $\sigma_{e-1}$, and a log excerpt $\xi(\Lambda)$, i.e., a few rows with their column names, or the log schema usually provided by the system manufacturer. The requested knowledge type $\kappa_e$ alternates between Semantic Insight and Surrogate Model across workflow runs, starting with an insight. To promote diversity, the prompt lists the hypotheses $\mathcal{H}$ proposed so far and asks for a different one, following the ``propose'' prompting strategy of \citet{tot}.

\textbf{Coder.} The Coder translates $h_e$ into the Diagnostic Artifact $A_e$, written for the log schema in $\xi(\Lambda)$ (challenge~$c$) and executed in a sandbox on the full buffer $\Lambda$ (challenge~$a$). For a Semantic Insight, $A_e$ prints the facts extracted from the logs. For a Surrogate Model, it fits a low-parameter regression, stores its weights, and defines a \texttt{predict} function with a usage example; the inference cost of such low-complexity models is negligible relative to the decision budget. If execution fails (e.g., syntax error or schema mismatch), the Coder regenerates $A_e$, up to $M_{\max}$ attempts.

\textbf{Merge.} Finally, an LLM rewrites the System Information to integrate the new knowledge, $\sigma_e = \textsc{Merge}(\sigma_{e-1}, K_e)$, including the signature and usage of any new tool (see Prompt~\ref{prompt:evolved_sys_info}). Merging, rather than appending, keeps $\sigma_e$ concise and coherent as knowledge accumulates.
Each hypothesis is thus tested by the generation loop: helpful knowledge is exploited by high-fitness offspring and propagates through the population, while unhelpful knowledge is simply ignored.

\subsection{Scheduling and cost}\label{sec:scheduling}

The Agentic Workflow runs at the start of each epoch until $E_{\max}$ runs have succeeded, i.e., their Diagnostic Artifacts executed; failed runs do not count toward this budget. Afterwards, the System Information is frozen, and the remaining epochs are purely evolutionary.
The agents are deliberately asymmetric. The Reasoner, which must reason about latent system dynamics from a log excerpt, uses a frontier reasoning model but is invoked only once per workflow run. The Coder, the Merge step, and the generation loop, which together make thousands of LLM calls, use a lightweight open-weight model. Consequently, the workflow adds less than $2\%$ wall-clock time over EoH at an equal generation budget (Sec.~\ref{sec:eval_cost}). Hyperparameters and models are listed in Appendix~\ref{sec:trace_conf}, and the effect of the LLM choice is analyzed in Appendix~\ref{app:llm_comparison}.

\section{Evaluation}\label{sec:eval1}

We evaluate whether \name{} (i) outperforms state-of-the-art AHD methods under the same generation budget, and how this advantage scales with the number of processors (Sec.~\ref{sec:cloud_computing}); (ii) carries over to a measurement-driven industrial system with a different objective (Sec.~\ref{sec:5g}); (iii) at what computational cost (Sec.~\ref{sec:eval_cost}); and (iv) whether its heuristics are more auditable (Sec.~\ref{sec:audit}).

\subsection{Evaluation Use Cases}

($i$) A synthetic \textit{cloud computing} use case that tests scalability in settings more complex than prior work~\citep{soualhia2015,bader2024lotaru,rosa2021computational}, e.g., 20 nodes of 5 types with 1-dimensional features and linear processing times in \citet{bader2024lotaru}. We consider \textbf{Scenario 1} ($N=10$) and \textbf{Scenario 2} ($N=50$), with all servers of different types and $d=5$ task features. Each heuristic is evaluated over a sequence of low, medium, and high loads to cover diverse operating conditions; it observes only the task features (not the queue states), and the drop penalty $\Gamma$ makes reliability dominate (Appendix~\ref{sec:exp_meth}).
($ii$) A measurement-driven \textit{5G vRAN} use case with $N=4$ heterogeneous processors and $d=3$ task features, whose processing times and energy consumption are non-linear, noisy functions of the features taken from real measurements~\citep{cloudric} (Sec.~\ref{sec:5g}).
The configuration of \name{} is given in Appendix~\ref{sec:trace_conf}.

\subsection{Baselines and Comparative Methods}

\textit{Rule-based policies:} \textbf{Random}, a load-agnostic uniform assignment (lower bound); \textbf{Oracle-Time}, which knows the processing times $t_{i,n}$ and greedily selects the fastest processor; and \textbf{Oracle-Queue}, which assigns each task to the processor with the lowest queue occupancy. Both oracles use privileged information. The 5G use case adds two industry policies, CPU-Only and HA-Only.

\textit{AHD frameworks:} \textbf{FunSearch}~\citep{funsearch}, the seminal LLM-based evolutionary program search (with a single-island database); \textbf{EoH}~\citep{eoh}, which adds structured evolutionary operators (crossover, mutation, simplification) through its five original prompts; \textbf{ReEvo}~\citep{reevo}, which adds ``verbal gradients'', i.e., natural-language critiques of the fitness history; and \textbf{HSEvo}~\citep{hsevo}, which combines genetic evolution and verbal reflection with harmony search to tune the numerical parameters of the best heuristic. All four rely exclusively on scalar fitness.

\textit{Protocol.} All AHD methods use the same budget of $3000$ generated heuristics ($E{=}20$ epochs of $I{=}150$ steps for \name{}) and the same open-weight LLM (Qwen3-30B-A3B); only \name{}'s Reasoner uses a frontier model (Appendix~\ref{sec:trace_conf}). 
Each method is run $10$ times; tables report the best run, and Fig.~\ref{fig:cost_dist} shows the distribution over runs.

\textit{Ablation.} Without the agentic workflow, \name{} is functionally equivalent to EoH, so their gap isolates the contribution of the log-derived knowledge. Moreover, runs in which the Coder fails to produce the surrogate degrade sharply, whereas replacing the Reasoner (Gemini, Claude, or GPT) leaves performance unchanged, as all three rediscover the same hypotheses (Appendix~\ref{app:llm_comparison}).

\subsection{Cloud Computing Use Case} \label{sec:cloud_computing}

\textbf{Performance comparison.}
Table~\ref{tab:best} reports the loss rate (fraction of dropped tasks), the mean delay of admitted tasks, and the cost $J$ (\eqref{eq:cost}).
\textit{Random} ignores processing heterogeneity and loses 70\% of the tasks in Scenario 1. The oracles expose the affinity--congestion trade-off: \textit{Oracle-Time} minimizes per-task latency but saturates fast processors, causing queueing and drops, whereas \textit{Oracle-Queue} eliminates drops in Scenario 2 but inflates delay by ignoring task--processor affinity.

A good policy must balance both, and \name{} synthesizes one: it achieves zero loss with the lowest delay in both scenarios and, in Scenario 2, it is the only AHD method with zero loss and the only method overall that attains it without inflating delay (0.44 vs.\ 17.32~ms for Oracle-Queue).
Among the AHD baselines, FunSearch and EoH tie in Scenario 1, and EoH's structured operators pay off in Scenario 2 (15.88 vs.\ 90.98). ReEvo's reflections help in Scenario 1, but without logs they can only guess the causes of poor performance from scalar costs, which becomes harder in Scenario 2. Likewise, HSEvo's parameter tuning yields a competitive policy in Scenario 1 but is less effective in Scenario 2, where the policy space grows with $N$.

\begin{figure}[t!]
\centering
\begin{subfigure}[b]{0.48\linewidth}
  \centering
  \includegraphics[width=0.9\linewidth]{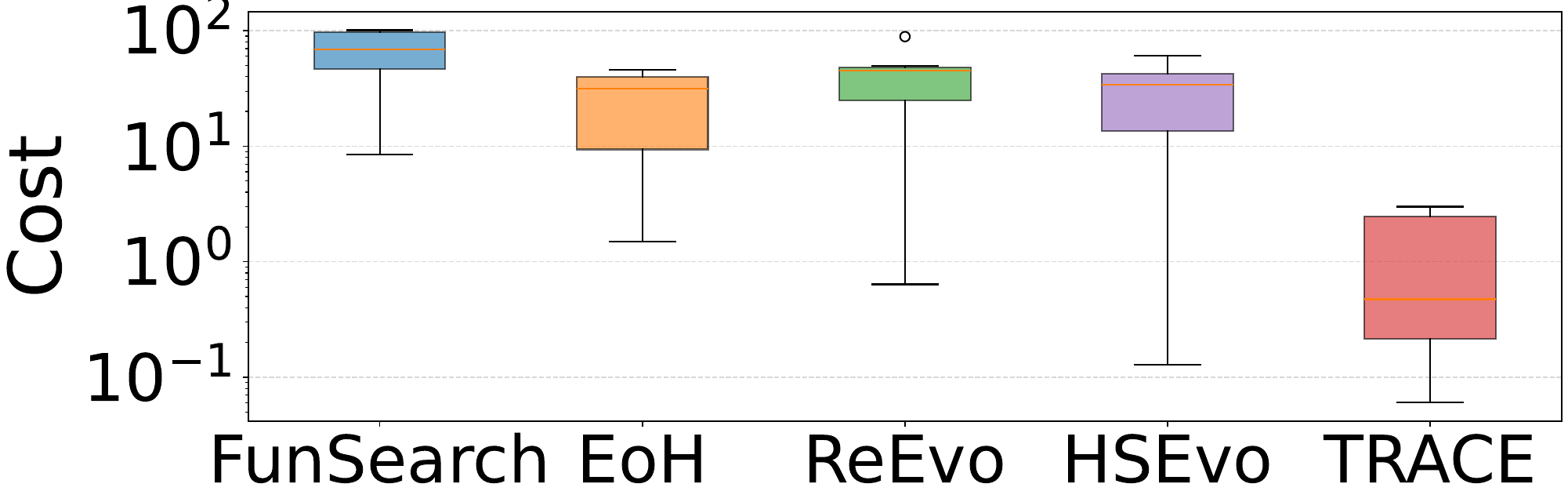}
  \label{fig:cost_dist_sc1}
\end{subfigure}\hfill
\begin{subfigure}[b]{0.48\linewidth}
  \centering
  \includegraphics[width=0.9\linewidth]{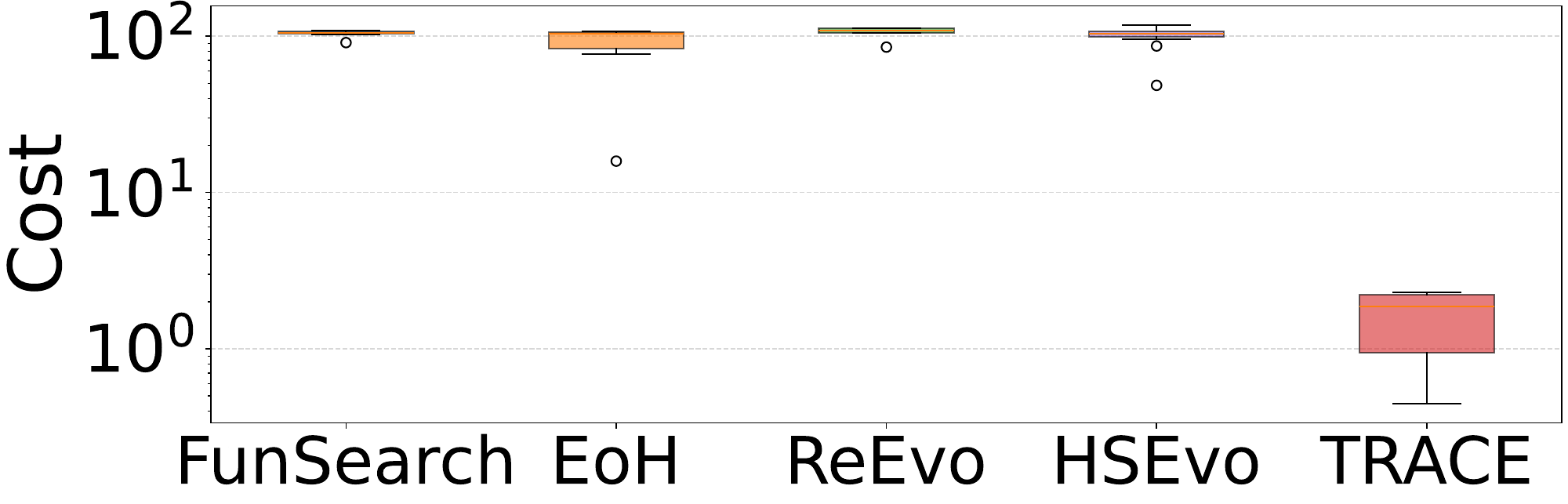}
  \label{fig:cost_dist_sc2}
\end{subfigure}
\caption{Cost $J$ (log scale, lower is better) over $10$ independent runs for Scenario 1 ($N{=}10$, left) and Scenario 2 ($N{=}50$, right).}
\label{fig:cost_dist}
\vspace{-1mm}
\end{figure}

The gap widens with scale: \name{}'s cost is about $2\times$ lower than that of the best AHD baseline in Scenario 1 (0.06 vs.\ 0.13, HSEvo) and about $36\times$ lower in Scenario 2 (0.44 vs.\ 15.88, EoH), as the space of assignment policies grows with the number of processors. This holds beyond the best run (Fig.~\ref{fig:cost_dist}): in Scenario 2, every run of \name{} beats every baseline run, and in Scenario 1 its median cost is more than an order of magnitude below the baselines' medians.

\textbf{Why does \name{} outperform the baselines?}
Its best heuristic in Scenario 1 (Algorithm~\ref{alg:ours_scenario_1}, Appendix~\ref{app:best_algo}) relies on physical properties of the system: it (i) calls the learned surrogate to estimate processing times, (ii) penalizes fast processors to prevent their saturation, (iii) uses the variance of processing times as a reliability measure, and (iv) balances the load with randomized top-$k$ selection. This knowledge comes from the agentic workflow, which, driven by the Diagnostic Hypotheses in Prompt~\ref{prompt:DH_example}, evolves the System Information prompt (Prompt~\ref{prompt:system_info}) into one that exposes a callable surrogate, per-processor speed statistics, and the dominant features of each processor (Prompt~\ref{prompt:evolved_sys_info}).
In contrast, the best baseline heuristics converge to \textit{deterministic hashing}, i.e., arbitrary arithmetic with no physical grounding, such as trigonometric scrambling (FunSearch), random-coefficient \texttt{tanh} mappings (EoH), modulo operations (ReEvo), and pseudo-random choices seeded by feature statistics (HSEvo). They perform worse and are also less auditable (Sec.~\ref{sec:audit}).

\begin{table}[t!]
\centering
\small
\begin{tabular}{l|ccc|ccc}
\hline
Strategy & \multicolumn{3}{c|}{Scenario 1 ($N{=}10$)} & \multicolumn{3}{c}{Scenario 2 ($N{=}50$)} \\
 & Loss & Delay (ms) & Cost ($J$) & Loss & Delay (ms) & Cost ($J$) \\ \hline
Random & 0.70 & 32.15 & 103.71 & 0.10 & 65.22 & 117.91 \\
Oracle-Time & 0.03 & 1.30 & 4.92 & 0.07 & 4.81 & 46.74 \\
Oracle-Queue & 0.62 & 15.76 & 90.23 & \textbf{0.00} & 17.32 & 17.32 \\  \hline
FunSearch & 0.05 & 2.43 & 8.50 & 0.08 & 49.39 & 90.98 \\
EoH & 0.05 & 2.43 & 8.50 & 0.01 & 11.51 & 15.88 \\
ReEvo & \textbf{0.00} & 0.64 & 0.64 & 0.07 & 46.22 & 85.16 \\
HSEvo & \textbf{0.00} & 0.13 & 0.13 & 0.07 & 9.38 & 48.50 \\
\textbf{\name{}} & \textbf{0.00} & \textbf{0.06} & \textbf{0.06} & \textbf{0.00} & \textbf{0.44} & \textbf{0.44} \\
\hline
\end{tabular}
\caption{Cloud-computing use case: loss rate, mean delay (ms), and cost $J$ (lower is better) of the fixed baselines and of the best of $10$ runs of each AHD method, over low, medium, and high loads.}\label{tab:best}
\end{table}

\subsection{Measurement-Driven Case Study: Energy-Efficient 5G vRAN} \label{sec:5g}

\textbf{Scenario.}
In 5G virtualized Radio Access Networks (vRAN), base station functions run on general-purpose edge clouds, and the computational bottleneck is uplink decoding. Each task is a Transport Block (TB) whose features, i.e., Transport Block Size, Modulation and Coding Scheme, and Signal-to-Noise Ratio, dictate its computational complexity. We simulate an edge cluster with $N=4$ heterogeneous processors (CPUs and hardware accelerators, HAs). The objective is \textit{energy efficiency} under a hard 1~ms deadline: TBs not decoded in time are dropped with penalty $\Gamma=10$ (\eqref{eq:cost_mobile}). Processing times and energy are queried from measurements of an industrial-grade vRAN testbed, and arrivals are replayed from traces of an operational mobile network~\citep{cloudric_dataset}; details are given in Appendix~\ref{app:mobile_motivation}.

\textbf{Result analysis.}
Table~\ref{table:mobile_best} reports the TB loss rate, mean processing delay, mean energy per TB, and cost.
The rule-based policies show how hard it is to meet hard real-time constraints on heterogeneous processors. \textit{Random} and \textit{Oracle-Queue} balance the load but lose about 10\% of the TBs, as computationally intensive tasks land on slow CPU cores. \textit{Oracle-Time} achieves 0\% loss by always choosing the fastest processor, but it is energy-agnostic: it uses high-power accelerators even for tasks that CPUs could finish in time, yielding the second-highest energy consumption.
Among current vRAN deployment policies, \textit{CPU-Only}~\citep{agora, concordia} has the lowest energy but a 27\% loss rate during traffic peaks, unacceptable for carrier-grade 5G, whereas \textit{HA-Only} (the industry standard) guarantees 0\% loss and minimal delay through over-provisioning: power-hungry accelerators process every task regardless of complexity, yielding the highest energy consumption.

Among AHD methods, we compare against EoH, the log-free counterpart of \name{} and the strongest AHD baseline in the larger cloud scenario, so that the gap measures the value of the logs. EoH finds a zero-loss policy with $35\%$ lower cost than the best fixed policy (0.58 vs.\ 0.89, Oracle-Time). \name{} also keeps 0\% loss while consuming the least energy among zero-loss policies: $64.9\%$ less than HA-Only and $25.4\%$ less than EoH (397.89 vs.\ 533.33~$\mu$J); its higher delay (0.36~ms) is well within the deadline, which is all the objective requires.
Since \name{} reduces to EoH without the agentic workflow, this gap is attributable to the knowledge extracted from the logs: a processing-time surrogate, the average energy of each processor, and the dominance of the transport block size on CPU processing time. The best heuristic combines them into a score that avoids processors whose predicted waiting plus processing time would exceed the deadline, and favours low-energy processors among the rest (Appendix~\ref{app:mobile}).

\begin{table}[t!]
\centering
\small
\begin{tabular}{l|cccc}
\hline
Strategy & Loss & Delay (ms) & Energy ($\mu$J) & Cost \\ \hline
Random & 0.10 & 0.30 & 670.01 & 1.58 \\
Oracle-Time & \textbf{0.00} & 0.35 & 884.22 & 0.89 \\
Oracle-Queue & 0.08 & 0.27 & 700.76 & 1.47 \\ \hline
CPU-Only & 0.27 & 0.59 & \textbf{170.46} & 2.79 \\
HA-Only & \textbf{0.00} & \textbf{0.21} & 1132.39 & 1.13 \\ \hline
EoH & \textbf{0.00} & 0.31 & 533.33 & 0.58 \\
\textbf{\name{}} & \textbf{0.00} & 0.36 & 397.89 & \textbf{0.40} \\
\hline
\end{tabular}
\caption{5G vRAN use case: TB loss rate, mean processing delay, mean energy per TB, and cost (\eqref{eq:cost_mobile}; lower is better) for fixed policies (top), industry standards (middle), and the best of $10$ runs of each AHD method (bottom).}
\label{table:mobile_best}
\vspace{-3mm}
\end{table}

\subsection{Computational overhead} \label{sec:eval_cost}
All methods share the same generation budget, and \name{}'s evolutionary backbone is functionally equivalent to EoH, so the runtime difference between them reduces to the agentic workflow. In an epoch in which it runs, the workflow adds $32.2$~s (Reasoner $22.5$~s, Coder $7.4$~s, sandbox $2.2$~s) to the $430.5$~s of heuristic generation and evaluation over the $I{=}150$ steps, i.e., $6.95\%$ of the epoch. The workflow runs only until $E_\text{max}{=}5$ successful executions, after which the System Information prompt is frozen and the remaining epochs are purely evolutionary. Over $E{=}20$ epochs, it thus accounts for $161$~s out of $2.44$~h, a $1.9\%$ wall-clock overhead relative to EoH, which does not grow with the number of processors since the workflow runs a fixed number of times. Full analysis in Appendix~\ref{app:cost}.

Token use is similarly asymmetric (Appendix~\ref{app:cost}): per run, the generation loop consumes about $1.5$M input and $0.8$--$2.2$M output tokens, and the Coder about $5$K input and $5$--$7$K output tokens, both on the local open-weight model used by all baselines. The Reasoner, the only component served by an external API, uses about $3$K tokens per run, three orders of magnitude fewer.

\subsection{Auditability Analysis}\label{sec:audit}

We evaluate auditability with an LLM-as-a-Judge panel of three models from distinct families (\texttt{Gemini~3.1~Pro}, \texttt{Claude~Opus~4.7}, \texttt{GPT-5.5}), scoring the best Scenario 1 heuristics from 1 to 5 on three rubrics: \textit{Semantic Grounding} (are the variables tied to physical system state?), \textit{Auditability} (how quickly can an operator locate and edit the logic governing processor saturation?), and \textit{Simulatability} (can the assignment be predicted by hand?). Each (judge, method, rubric) triple is queried five times at temperature $0.7$ (details in Appendix~\ref{app:audit}).

\name{} is the only method that scores above 1.3 on Semantic Grounding (3.33 vs.\ 1.00--1.27; Table~\ref{tab:audit_main}), as its branching logic and penalty factors are tied to predicted processing times and thresholds, whereas the baselines rely on abstract mathematical transformations. On Auditability, it scores 4.67 against 1.33--2.00, since no baseline exposes a per-processor threshold that an operator could tighten. Every judge ranks \name{} first on both rubrics, which mitigates concerns about judge-specific bias, including from the Gemini judge, whose family also serves as \name{}'s Reasoner. With a single heuristic per method, the scores should be read as indicative.

Simulatability exposes a trade-off: \name{}'s policy calls the learned \texttt{predict()} surrogate and cannot be evaluated by hand (1.00), whereas ReEvo's three-line modulo rule can (5.00) but encodes no system property. The surrogate itself is a low-order regression over the task features that can be inspected separately, whereas hashing rules admit no such decomposition.

\begin{table}[t!]
\centering
\small
\begin{tabular}{l|ccc|c}
\hline
Strategy & Semantic Grounding & Auditability & Simulatability & Mean \\ \hline
FunSearch      & 1.00 & 1.47 & 2.00          & 1.49 \\
EoH            & 1.00 & 1.33 & 4.47          & 2.27 \\
ReEvo          & 1.00 & 2.00 & \textbf{5.00} & 2.67 \\
HSEvo          & 1.27 & 1.33 & 2.27          & 1.62 \\
\textbf{\name{}} & \textbf{3.33} & \textbf{4.67} & 1.00 & \textbf{3.00} \\
\hline
\end{tabular}
\caption{LLM-as-a-Judge auditability scores for the best heuristics in Scenario~1, averaged over a three-judge panel and five queries per judge. Per-judge scores and rubric definitions are in Appendix~\ref{app:audit}.}
\label{tab:audit_main}
\end{table}

\section{Related Work}\label{sec:related}

\textbf{Dynamic resource assignment.}
Resource assignment is commonly formulated as the Dynamic Job-Shop Scheduling Problem (DJSSP), addressed with hand-crafted heuristics, DRL, and graph-based policies~\citep{djssp_1, djssp_2}. LLM-based AHD has also been applied to job-shop variants: LLMMA couples an LLM with a memetic algorithm for lot-streaming hybrid job shops~\citep{li2025llmma}, SeEvo performs population self-evolution for the DJSSP~\citep{huang2024seevo}, and recent works target dynamic flexible job shops with evolutionary textual gradients~\citep{ijpr_dfjsp} and job shops with preventive maintenance~\citep{dgevo}. These works typically assume deterministic and known processing times and infinite buffers, which do not hold in the streaming, finite-buffer systems targeted in this work (Sec.~\ref{sec:prob_form}).

\textbf{LLM-based Automatic Heuristic Design.}
Early genetic-programming approaches~\citep{genetic_programming} evolved tree-structured heuristics within rigid search spaces; LLMs have since enabled the evolution of arbitrary code~\citep{llmea_survey}. FunSearch~\citep{funsearch} established the LLM-driven evolutionary loop, EoH~\citep{eoh} introduced structured evolutionary operators, and ReEvo~\citep{reevo} added ``verbal gradients'' derived from the performance history. Subsequent works employ harmony search~\citep{hsevo}, reinforcement learning~\citep{calm}, Monte Carlo Tree Search~\citep{mcts}, and niche-based search over language spaces~\citep{hu2025partevo}, while AlphaEvolve~\citep{novikov2025alphaevolve} extends the paradigm to broad scientific and algorithmic discovery. Recently, MoH meta-learns the heuristic optimizer itself~\citep{moh2026}, EoH-S evolves complementary sets of heuristics~\citep{eohs2026}, and PathWise plans over the space of heuristics with a world model~\citep{pathwise2026}. They are evaluated on combinatorial and black-box optimization benchmarks~\citep{cobench, heurigym, blade}. Such advances change \textit{how} the search proceeds, whereas \name{} changes \textit{what the LLM observes}; its Agentic Workflow is thus orthogonal to the choice of evolutionary backbone.
Regarding feedback, ReEvo and HSEvo verbalize reflections derived from the fitness history~\citep{reevo, hsevo}, DHEvo co-evolves training data with primal-gap feedback~\citep{zhang2025dhevo}, and MLES incorporates multimodal feedback through pre-engineered visual pipelines~\citep{hu2025mles}. In all these methods, the LLM receives either the fitness, possibly verbalized, or feedback whose content is fixed by the framework designer. To our knowledge, \name{} is the first AHD framework in which an agent decides \textit{what} to extract from the execution logs of the target system and writes the code that extracts it, including executable surrogate models that the evolved heuristics can call.

\textbf{LLM-based design for networked and computing systems.}
Beyond AlphaEvolve's improved production data-center scheduler~\citep{novikov2025alphaevolve}, AHD has evolved 5G LDPC decoding functions~\citep{iccspa_ldpc, ahd_netops} and routing heuristics that are re-synthesized upon traffic-regime shifts and repaired from verifier traces~\citep{rass_evo}, and LLMs have guided virtual network embedding~\citep{flexvne} and vehicular edge computing heuristics~\citep{panorama}. None of them derives knowledge about the unknown system dynamics from execution logs.

\section{Limitations}\label{sec:limitations}

First, \name{}'s performance depends on agents capable of deducing latent dynamics and writing valid extraction code, and on execution logs with informative columns. When these are lacking, as in the Llama-3.1-70B runs where the Coder failed to generate valid code to parse the logs and train a surrogate (Appendix~\ref{app:llm_comparison}), or when logs lack the relevant fields, the agentic workflow yields no useful knowledge and \name{} degrades gracefully to its scalar-feedback backbone (i.e., EoH), so its worst-case performance is lower-bounded by standard AHD.

Second, the one-time discovery cost (roughly 2.5 wall-clock hours; Appendix~\ref{app:cost}) is amortized over the deployment lifetime of the heuristic but non-negligible if hardware changes frequently. Hardware updates, however, need not restart from scratch: seeding the evolutionary loop with previous high-performing heuristics and new platform logs lets the agentic workflow re-derive all hardware-dependent quantities (e.g., per-processor energy profiles, surrogate models) and adapt existing heuristics instead of rediscovering them, substantially reducing compute.

Third, for highly complex dynamics, the surrogate regressors needed may exceed the latency budget. However, low-order regressions give accurate approximations in our benchmarks, which are representative of real-world resource-assignment systems~\citep{soualhia2015,bader2024lotaru, cloudric, cloudric_dataset}.

\section{Conclusion}

We presented \name{}, which extends LLM-based Automatic Heuristic Design beyond scalar feedback: an agentic workflow distills system execution logs into insights and executable surrogate models that guide heuristic evolution. \name{} outperforms state-of-the-art AHD methods on cloud and 5G vRAN scenarios, e.g., cutting energy by 65\% relative to the industry standard without losing reliability, and yields more auditable heuristics at under 2\% overhead. We show that system logs are thus a valuable, largely untapped source of feedback for AHD.

\newpage

\bibliography{references}

@inproceedings{resource_microsoft,
  title={Resource management with deep reinforcement learning},
  author={Mao, Hongzi and Alizadeh, Mohammad and Menache, Ishai and Kandula, Srikanth},
  booktitle={Proceedings of the 15th ACM workshop on hot topics in networks},
  pages={50--56},
  year={2016}
}

@article{tot,
  title={Tree of thoughts: Deliberate problem solving with large language models},
  author={Yao, Shunyu and Yu, Dian and Zhao, Jeffrey and Shafran, Izhak and Griffiths, Tom and Cao, Yuan and Narasimhan, Karthik},
  journal={Advances in neural information processing systems},
  volume={36},
  pages={11809--11822},
  year={2023}
}

@article{reevo,
  title={Reevo: Large language models as hyper-heuristics with reflective evolution},
  author={Ye, Haoran and Wang, Jiarui and Cao, Zhiguang and Berto, Federico and Hua, Chuanbo and Kim, Haeyeon and Park, Jinkyoo and Song, Guojie},
  journal={Advances in neural information processing systems},
  volume={37},
  pages={43571--43608},
  year={2024}
}

@inproceedings{agora,
  title={Agora: Real-time massive MIMO baseband processing in software},
  author={Ding, Jian and Doost-Mohammady, Rahman and Kalia, Anuj and Zhong, Lin},
  booktitle={Proceedings of the 16th international conference on emerging networking experiments and technologies},
  pages={232--244},
  year={2020}
}

@inproceedings{concordia,
  title={Concordia: Teaching the 5G vRAN to share compute},
  author={Foukas, Xenofon and Radunovic, Bozidar},
  booktitle={Proceedings of the 2021 ACM SIGCOMM 2021 Conference},
  pages={580--596},
  year={2021}
}

@misc{cloudric_dataset,
  title={Latency and energy characterization of 5G LDPC FEC Decoding on CPU and GPU},
  author={Lo Schiavo, Leonardo and Garcia-Aviles, Gines and Garcia-Saavedra, Andres and Gramaglia, Marco and Fiore, Marco and Banchs, Albert and Costa-Perez, Xavier},
  journal={Latency and energy characterization of 5G LDPC FEC Decoding on CPU and GPU},
  year={2024},
  publisher={Zenodo},
  doi={10.5281/zenodo.10691661},
  url={https://doi.org/10.5281/zenodo.10691661},
}

@article{eoh,
  title={Evolution of heuristics: Towards efficient automatic algorithm design using large language model},
  author={Liu, Fei and Tong, Xialiang and Yuan, Mingxuan and Lin, Xi and Luo, Fu and Wang, Zhenkun and Lu, Zhichao and Zhang, Qingfu},
  journal={arXiv preprint arXiv:2401.02051},
  year={2024}
}

@article{funsearch,
  title={Mathematical discoveries from program search with large language models},
  author={Romera-Paredes, Bernardino and Barekatain, Mohammadamin and Novikov, Alexander and Balog, Matej and Kumar, M Pawan and Dupont, Emilien and Ruiz, Francisco JR and Ellenberg, Jordan S and Wang, Pengming and Fawzi, Omar and others},
  journal={Nature},
  volume={625},
  number={7995},
  pages={468--475},
  year={2024},
  publisher={Nature Publishing Group UK London}
}

@article{cloud_survey_1,
  title={A survey study on task scheduling schemes for workflow executions in cloud computing environment: classification and challenges},
  author={Hosseini Shirvani, Mirsaeid},
  journal={The Journal of Supercomputing},
  volume={80},
  number={7},
  pages={9384--9437},
  year={2024},
  publisher={Springer}
}

@article{cloud_suvery_2,
  title={Cluster resource scheduling in cloud computing: literature review and research challenges},
  author={Khallouli, Wael and Huang, Jingwei},
  journal={The Journal of supercomputing},
  volume={78},
  number={5},
  pages={6898--6943},
  year={2022},
  publisher={Springer}
}

@inproceedings{cloudric,
  title={Cloudric: Open radio access network (o-ran) virtualization with shared heterogeneous computing},
  author={Schiavo, Leonardo Lo and Garcia-Aviles, Gines and Garcia-Saavedra, Andres and Gramaglia, Marco and Fiore, Marco and Banchs, Albert and Costa-Perez, Xavier},
  booktitle={Proceedings of the 30th Annual International Conference on Mobile Computing and Networking},
  pages={558--572},
  year={2024}
}

@article{aslan2024fair,
  title={Fair Resource Allocation in Virtualized O-RAN Platforms},
  author={Aslan, Fatih and Iosifidis, George and Ayala-Romero, Jose A and Garcia-Saavedra, Andres and Costa-Perez, Xavier},
  journal={Proceedings of the ACM on Measurement and Analysis of Computing Systems},
  volume={8},
  number={1},
  pages={1--34},
  year={2024},
  publisher={ACM New York, NY, USA}
}

@article{djssp_1,
  title={Dynamic job-shop scheduling problems using graph neural network and deep reinforcement learning},
  author={Liu, Chien-Liang and Huang, Tzu-Hsuan},
  journal={IEEE transactions on systems, man, and cybernetics: systems},
  volume={53},
  number={11},
  pages={6836--6848},
  year={2023},
  publisher={IEEE}
}

@article{djssp_2,
  title={Dynamic job shop scheduling based on deep reinforcement learning for multi-agent manufacturing systems},
  author={Zhang, Yi and Zhu, Haihua and Tang, Dunbing and Zhou, Tong and Gui, Yong},
  journal={Robotics and Computer-Integrated Manufacturing},
  volume={78},
  pages={102412},
  year={2022},
  publisher={Elsevier}
}

@article{task_assignment,
  title={Task placement and resource allocation for edge machine learning: A gnn-based multi-agent reinforcement learning paradigm},
  author={Li, Yihong and Zhang, Xiaoxi and Zeng, Tianyu and Duan, Jingpu and Wu, Chuan and Wu, Di and Chen, Xu},
  journal={IEEE Transactions on Parallel and Distributed Systems},
  volume={34},
  number={12},
  pages={3073--3089},
  year={2023},
  publisher={IEEE}
}

@article{calm,
  title={CALM: Co-evolution of algorithms and language model for automatic heuristic design},
  author={Huang, Ziyao and Wu, Weiwei and Wu, Kui and Wang, Jianping and Lee, Wei-Bin},
  journal={arXiv preprint arXiv:2505.12285},
  year={2025}
}

@inproceedings{hsevo,
  title={Hsevo: Elevating automatic heuristic design with diversity-driven harmony search and genetic algorithm using llms},
  author={Dat, Pham Vu Tuan and Doan, Long and Binh, Huynh Thi Thanh},
  booktitle={Proceedings of the AAAI Conference on Artificial Intelligence},
  year={2025},
  volume={39},
  number={25},
  pages={26931--26938},
  series={AAAI '25}
}

@inproceedings{mcts,
  title={Monte Carlo Tree Search for Comprehensive Exploration in LLM-Based Automatic Heuristic Design},
  author={Zheng, Zhi and Xie, Zhuoliang and Wang, Zhenkun and Hooi, Bryan},
  booktitle={Forty-second International Conference on Machine Learning},
  year={2025}
}

@article{genetic_programming,
  title={Survey on genetic programming and machine learning techniques for heuristic design in job shop scheduling},
  author={Zhang, Fangfang and Mei, Yi and Nguyen, Su and Zhang, Mengjie},
  journal={IEEE Transactions on Evolutionary Computation},
  volume={28},
  number={1},
  pages={147--167},
  year={2023},
}

@article{li2025llmma,
  title={LLM-assisted automatic memetic algorithm for lot-streaming hybrid job shop scheduling with variable sublots},
  author={Li, Rui and Wang, Ling and Sang, Hongyan and Yao, Lizhong and Pan, Lijun},
  journal={IEEE Transactions on Evolutionary Computation},
  year={2025},
  publisher={IEEE}
}

@article{huang2024seevo,
  title={Automatic programming via large language models with population self-evolution for dynamic fuzzy job shop scheduling problem},
  author={Huang, Jin and Liu, Qihao and Li, Xinyu and Gao, Liang and Teng, Yue},
  journal={IEEE Transactions on Fuzzy Systems},
  year={2026},
  publisher={IEEE}
}

@article{novikov2025alphaevolve,
  title={Alphaevolve: A coding agent for scientific and algorithmic discovery},
  author={Novikov, Alexander and V{\~u}, Ng{\^a}n and Eisenberger, Marvin and Dupont, Emilien and Huang, Po-Sen and Wagner, Adam Zsolt and Shirobokov, Sergey and Kozlovskii, Borislav and Ruiz, Francisco JR and Mehrabian, Abbas and others},
  journal={arXiv preprint arXiv:2506.13131},
  year={2025}
}

@inproceedings{hu2025partevo,
  title={Partition to evolve: Niching-enhanced evolution with llms for automated algorithm discovery},
  author={Hu, Qinglong and Zhang, Qingfu},
  booktitle={The Thirty-ninth Annual Conference on Neural Information Processing Systems},
  year={2025}
}

@article{zhang2025dhevo,
  title={DHEvo: Data-Algorithm Based Heuristic Evolution for Generalizable MILP Solving},
  author={Zhang, Zhihao and Li, Siyuan and Li, Chenxi and Liu, Feifan and Chen, Mengjing and Li, Kai and Zhong, Tao and An, Bo and Liu, Peng},
  journal={arXiv preprint arXiv:2507.15615},
  year={2025}
}

@article{hu2025mles,
  title={Multimodal LLM-assisted Evolutionary Search for Programmatic Control Policies},
  author={Hu, Qinglong and Tong, Xialiang and Yuan, Mingxuan and Liu, Fei and Lu, Zhichao and Zhang, Qingfu},
  journal={arXiv preprint arXiv:2508.05433},
  year={2025}
}

@inproceedings{soualhia2015,
  title={Predicting scheduling failures in the cloud: A case study with google clusters and hadoop on amazon EMR},
  author={Soualhia, Mbarka and Khomh, Foutse and Tahar, Sofiene},
  booktitle={2015 IEEE 17th International Conference on High Performance Computing and Communications, 2015 IEEE 7th International Symposium on Cyberspace Safety and Security, and 2015 IEEE 12th International Conference on Embedded Software and Systems},
  pages={58--65},
  year={2015},
  organization={IEEE}
}

@article{bader2024lotaru,
  title={Lotaru: Locally predicting workflow task runtimes for resource management on heterogeneous infrastructures},
  author={Bader, Jonathan and Lehmann, Fabian and Thamsen, Lauritz and Leser, Ulf and Kao, Odej},
  journal={Future Generation Computer Systems},
  volume={150},
  pages={171--185},
  year={2024},
  publisher={Elsevier}
}

@inproceedings{heurigym,
  title={HeuriGym: An Agentic Benchmark for LLM-Crafted Heuristics in Combinatorial Optimization},
  author={Chen, Hongzheng and Wang, Yingheng and Cai, Yaohui and Hu, Hins and Li, Jiajie and Huang, Shirley and Deng, Chenhui and Liang, Rongjian and Kong, Shufeng and Ren, Haoxing and others},
  booktitle={The 5th Workshop on Mathematical Reasoning and AI at NeurIPS 2025},
  year={2025}
}

@inproceedings{cobench,
  title={Co-bench: Benchmarking language model agents in algorithm search for combinatorial optimization},
  author={Sun, Weiwei and Feng, Shengyu and Li, Shanda and Yang, Yiming},
  booktitle={Proceedings of the AAAI Conference on Artificial Intelligence},
  volume={40},
  number={39},
  pages={33126--33134},
  year={2026}
}

@article{rosa2021computational,
  title={Computational resource and cost prediction service for scientific workflows in federated clouds},
  author={Rosa, Michel JF and Ralha, C{\'e}lia Ghedini and Holanda, Maristela and Araujo, Aleteia PF},
  journal={Future Generation Computer Systems},
  volume={125},
  pages={844--858},
  year={2021},
  publisher={Elsevier}
}

@inproceedings{moh2026,
  title={Generalizable Heuristic Generation Through {LLMs} with Meta-Optimization},
  author={Shi, Yiding and Zhou, Jianan and Song, Wen and Bi, Jieyi and Wu, Yaoxin and Cao, Zhiguang and Zhang, Jie},
  booktitle={International Conference on Learning Representations (ICLR)},
  year={2026}
}

@inproceedings{eohs2026,
  title={{EoH-S}: Evolution of Heuristic Set Using {LLMs} for Automated Heuristic Design},
  author={Liu, Fei and Liu, Yilu and Zhang, Qingfu and Tong, Xialiang and Yuan, Mingxuan},
  booktitle={Proceedings of the AAAI Conference on Artificial Intelligence},
  volume={40},
  number={43},
  pages={37090--37098},
  year={2026},
  doi={10.1609/aaai.v40i43.41038}
}

@inproceedings{pathwise2026,
  title={{PathWise}: Planning through World Model for Automated Heuristic Design via Self-Evolving {LLMs}},
  author={Gungordu, Oguzhan and Xiong, Siheng and Fekri, Faramarz},
  booktitle={International Conference on Machine Learning (ICML)},
  year={2026}
}

@inproceedings{iccspa_ldpc,
  title={Exploring the Viability of Automated Heuristic Design for {5G} {LDPC} Decoding},
  author={Namvar, Reza and Gallego, Jos{\'e} and Banchs, Albert and Chatzieleftheriou, Livia Elena and Ayala-Romero, Jose A. and Garcia-Saavedra, Andr{\'e}s and Fiore, Marco},
  booktitle={International Conference on Communications, Signal Processing, and their Applications (ICCSPA)},
  pages={1--6},
  year={2026},
  doi={10.1109/ICCSPA69228.2026.11600860}
}

@article{ahd_netops,
  title={Automated Heuristic Design for Network Operations},
  author={Namvar, Reza and Gallego, Jos{\'e} and Ayala-Romero, Jose A. and Chatzieleftheriou, Livia Elena and Garcia-Saavedra, Andr{\'e}s and Banchs, Albert and Fiore, Marco},
  journal={arXiv preprint arXiv:2605.28197},
  year={2026}
}

@article{rass_evo,
  title={{LLM}-Driven Regime-Adaptive Strategy Synthesis for Polymorphic Network Routing},
  author={Guo, Shuhan and Bai, Yudong and Huangfu, Wei and Zheng, Quan and Yao, Quanming},
  journal={Neural Networks},
  volume={204},
  pages={109269},
  year={2026},
  doi={10.1016/j.neunet.2026.109269}
}

@article{dgevo,
  title={{DGEvo}: Automatic Heuristic Design via Dual-Feature-Guided Evolution with Large Language Models for Job Shop Scheduling Problem with Fixed Preventive Maintenance},
  author={Liu, Lili and Zhang, Jingfei and Cheng, Weiyao and Meng, Leilei and Gao, Fangzheng},
  journal={Swarm and Evolutionary Computation},
  volume={107},
  pages={102477},
  year={2026},
  doi={10.1016/j.swevo.2026.102477}
}

@article{ijpr_dfjsp,
  title={Automated Design of Scheduling Heuristics for Dynamic Flexible Job Shop Scheduling with Evolutionary Textual Gradients},
  author={Huang, Jin and Zhang, Hao and Liu, Qihao and Li, Xinyu and Gao, Liang},
  journal={International Journal of Production Research},
  pages={1--22},
  year={2026},
  doi={10.1080/00207543.2026.2696443}
}

@article{flexvne,
  title={{FlexVNE}: A {LLM}-Powered Flexible Framework for Virtual Network Embedding in Cloud-Network Convergence},
  author={Pan, Fan and Tian, Le and Li, Dan and Cui, Pengshuai and Wang, Yu and Song, Xuanyan and Li, Menglong and Pei, Xinglong and Hu, Yuxiang and Guo, Xiaobo},
  journal={IEEE Transactions on Mobile Computing},
  volume={25},
  pages={18357--18373},
  year={2026},
  doi={10.1109/TMC.2026.3700558}
}

@inproceedings{panorama,
  title={Panorama: {LLM}-Guided Heuristics for Digital Twin-Empowered Vehicular Edge Computing},
  author={Huang, Ziyao and Wu, Kui and Wu, Weiwei and Qi, Xiangtong and Wang, Jianping and Wu, Jen-Ming},
  booktitle={IEEE International Conference on Distributed Computing Systems (ICDCS)},
  pages={1334--1344},
  year={2026},
  doi={10.1109/2575-8411.2026.00130}
}

@inproceedings{blade,
  title={{BLADE}: Benchmark Suite for {LLM}-Driven Automated Design and Evolution of Iterative Optimisation Heuristics},
  author={van Stein, Niki and Kononova, Anna V. and Yin, Haoran and B{\"a}ck, Thomas},
  booktitle={Proceedings of the Genetic and Evolutionary Computation Conference Companion (GECCO)},
  pages={2336--2344},
  year={2025},
  doi={10.1145/3712255.3734347}
}

@article{llmea_survey,
  title={Exploring Automated Algorithm Design Synergizing Large Language Models and Evolutionary Algorithms: Survey and Insights},
  author={Yu, He and Liu, Jing},
  journal={Evolutionary Computation},
  volume={34},
  number={2},
  pages={167--194},
  year={2026},
  doi={10.1162/EVCO.a.370}
}
\bibliographystyle{iclr2027_conference}

\appendix 
\newpage
\section{TRACE's algorithm}
\begin{algorithm}[h]
\caption{\name{}}\label{alg:trace}
\begin{algorithmic}[1]
\Require problem description $\Pi$, initial System Information $\sigma_0$, budgets $E$, $I$, $E_{\max}$, $M_{\max}$
\State $\Lambda \gets$ log of a random assignment policy; \; $\mathcal{D} \gets$ initial population (EoH initialization with $\sigma_0$)
\State $\mathcal{K} \gets \emptyset$; \; $\mathcal{H} \gets \emptyset$; \; $w \gets 0$ \Comment{$w$: successful workflow runs}
\For{$e = 1,\dots,E$}
  \State $\sigma_e \gets \sigma_{e-1}$
  \If{$w < E_{\max}$} \Comment{Agentic Workflow}
    \State $\kappa_e \gets$ \textsc{Insight} \textbf{if} $w$ is even \textbf{else} \textsc{Surrogate}
    \State $h_e \gets \textsc{Reasoner}(\Pi, \sigma_{e-1}, \xi(\Lambda), \kappa_e, \mathcal{H})$; \; $\mathcal{H} \gets \mathcal{H} \cup \{h_e\}$
    \For{$a = 1,\dots,M_{\max}$}
      \State $A_e \gets \textsc{Coder}(h_e, \xi(\Lambda))$; \; execute $A_e$ on $\Lambda$ in a sandbox
      \If{execution succeeds}
        \State $K_e \gets$ output of $A_e$; \; $\mathcal{K} \gets \mathcal{K} \cup \mathrm{tools}(K_e)$
        \State $\sigma_e \gets \textsc{Merge}(\sigma_{e-1}, K_e)$; \; $w \gets w+1$; \; \textbf{break}
      \EndIf
    \EndFor
  \EndIf
  \For{$j = 1,\dots,I$} \Comment{generation loop}
    \State $g \gets \textsc{LLM}(\Pi, \sigma_e, \textsc{Parents}_{m}(\mathcal{D}), \textsc{Operator})$
    \State evaluate $g \in \mathcal{G}(\mathcal{K})$, obtaining $J(g)$ and $\Lambda(g)$
    \If{$g$ is valid}
      \State update $\mathcal{D}$ with $(g, J(g))$; \; append $\Lambda(g)$ to $\Lambda$, discarding the oldest records beyond capacity
    \EndIf
  \EndFor
\EndFor
\State \Return $\arg\min_{g \in \mathcal{D}} J(g)$
\end{algorithmic}
\end{algorithm}

\section{Experimental Methodology: Cloud Computing use case}\label{sec:exp_meth}
\textbf{Simulation Environment.} We consider a task feature space of dimensionality $d=5$, where each task $i$ is represented by a feature vector $\displaystyle \vx_i \in [0, 1]^d$. The system comprises $N$ heterogeneous processors. To model hardware heterogeneity, each processor $n$ is characterized by a hidden capability vector $\displaystyle \vh_n \in [0, 1]^d$.
Following the specialized literature, we model the processing time $t_{i,n}$ as the alignment between the task requirements and the processor's capabilities, i.e., $t_{i,n} = \displaystyle \vx_i^\top \displaystyle \vh_n$~\citep{soualhia2015,bader2024lotaru, rosa2021computational}; non-linear dynamics are stress-tested in the 5G case study (Sec.~\ref{sec:5g}).
In this formulation, some processors can be specialized in one or several features, which implies a small value for the corresponding features. 
Each processor $n$ maintains a finite First-In-First-Out (FIFO) queue with capacity $L_n$, drawn from $[5, 15]$. If a task is assigned to a processor where the queue is full, the task is dropped immediately.

\textbf{Task generation.} Tasks are generated in discrete-time slots of 1 ms, with the number of tasks per slot drawn from a uniform distribution $U[0, \lambda]$. We design the distribution such that 90\% of tasks are ``specialized" (having high magnitude in specific dimensions and low noise elsewhere), while the remaining 10\% have uniformly random features.
Note that the heuristic observes only the task vector $\displaystyle \vx_i$, lacking access to the hidden processor vectors $\{\displaystyle \vh_n\}_{n=1}^N$ and the processors' instantaneous queue lengths.
The system generates a detailed post-hoc CSV execution log. Each discrete task event records: (i) \textit{Arrival context:} Timestamp and feature vector $\displaystyle \vx_i$; (ii) \textit{Decision:} The selected processor $n$; (iii) \textit{Outcome:} Processing latency $t_{i,n}$ (if admitted), binary drop status, and arrival queue states.

\textbf{Experimental Protocol.} We define two evaluation scenarios comprising $N=10$ and $N=50$ processors. For each scenario, we evaluate the generated algorithms over a simulation window of 3,000 time steps (3 seconds). Each heuristic is evaluated over three load levels in sequence: Low ($\lambda=2$), Medium ($\lambda=5$), and High ($\lambda=10$), so that its cost reflects a variety of operating conditions.
Performance is quantified using three metrics: (i) \textit{Loss}, the probability of task drop; (ii) \textit{Delay}, the mean latency for admitted tasks; and (iii) the \textit{Cost} as defined in Eq.~\ref{eq:cost}. We calibrate the drop penalty $\Gamma$ to enforce a high priority on reliability, setting $\Gamma=135$ for the 10-processor scenario and $\Gamma=600$ for the 50-processor scenario.

\section{TRACE configuration}\label{sec:trace_conf}

We configure \name{} with $I=150$, $E=20$, $M_\text{max}=10$, and $E_\text{max}=5$. 
All the prompts are reported in Appendix~\ref{app:prompts} (see prompt~\ref{prompt:prob_descr}-\ref{prompt:DA_3_insight}). A system log excerpt of the cloud computing and 5G use cases provided in Prompt~\ref{prompt:log_example} and \ref{prompt:log_example_5G}, respectively.

\name{} deliberately decouples reasoning from code generation. The Reasoner Agent (Sec.~\ref{sec:method}) requires deductive reasoning over latent system dynamics and is instantiated with a frontier reasoning model (\texttt{gemini-3-pro-preview}). All other components (Coder Agent and the heuristic-generation loop) use the open-weight \texttt{Qwen3-30B-A3B-Instruct-2507-FP8}, exactly matching the LLM used by all baselines (FunSearch, EoH, ReEvo, HSEvo). The asymmetric allocation is a deliberate efficiency contribution: the Reasoner is invoked only $E_{\max}{=}5$ times per run, whereas the Coder/Evolution loop performs thousands of generations. We quantify this cost asymmetry in Sec.~\ref{sec:eval_cost} and Appendix~\ref{app:cost}. We also evaluate the performance contribution of different LLM backbones in Appendix~\ref{app:llm_comparison}.

All methods are run under the same number of generations: \name{} for $E{=}20$ epochs of $I{=}150$ generations, and the AHD methods (FunSearch, EoH, ReEvo, HSEvo) for the equivalent $3000$ generations. 
\newpage
\section{Computational Cost Analysis}\label{app:cost}

This appendix profiles the runtime and token consumption of \name{} on the synthetic benchmark of Sec.~\ref{sec:eval1}, and contrasts it with standard AHD baselines (e.g., EoH).

\paragraph{Token economy.}
Table~\ref{tab:cost_tokens} reports the per-component profile. The only component that consumes external API tokens is the Reasoner (i.e., Google API), which uses fewer than 600 input tokens and 30 output tokens per call. The heavy-lifting components (Coder, Heuristic generation, Sandbox/Heuristic evaluation) are completely offloaded to a local non-reasoning 30B LLM and CPU execution.

\begin{table}[h!]
\centering
\small
\begin{tabular}{lllll}
\toprule
\textbf{Component} & \textbf{Hardware / Service} & \textbf{Tokens (in)} & \textbf{Tokens (out)} & \textbf{Time (s)} \\
\midrule
Reasoner (Hypothesis gen.) & Google API & 454--580 & 21--25 & 14--31 \\
Coder (Artifact generation) & Local GPU A100 & 980--995 & 1050--1350 & 7.3--7.8 \\
Sandbox execution & Local CPU Intel i7 & -- & -- & 0.1--3.85 \\
Heuristic generation & Local GPU A100 & 479--495 & 262--738 & 1.7--3.5 \\
Heuristic evaluation & Local CPU Intel i7 & -- & -- & 0.82--1.3 \\
\bottomrule
\end{tabular}
\caption{Per-component computational profile. Token consumption and execution times are reported as the 5th and 95th percentiles over 10 independent runs.}
\label{tab:cost_tokens}
\end{table}

\paragraph{Per-epoch wall-clock decomposition.}
Table~\ref{tab:cost_epoch} breaks down the wall-clock time of a single \name{} epoch ($I=150$ generation steps). The standard AHD steps (heuristic generation and evaluation) account for 430.5~s (93.05\%), which corresponds to the cost of running a standard AHD baselines (e.g., EoH). \name{}'s novel Agentic Workflow (Reasoner + Coder + Sandbox) requires only $\sim$32~s on average, an overhead of 6.95\% relative to the total epoch time. Even when the Coder requires multiple retries (up to $M_{\max}$), the 7.4~s cost per retry remains marginal; in practice, with Qwen3-30B we rarely observe retries.
Note that Table~\ref{tab:cost_epoch} profiles an epoch \textit{in which the agentic workflow is executed}. Since the workflow is invoked only $E_\text{max}$ times (Sec.~\ref{sec:method}), the remaining epochs are purely evolutionary and cost $430.5$~s each.
That is, over a complete run of 20 epochs, the total wall-clock time is $5 \times 462.65 + 15 \times 430.5 = 8771$~s ($\approx$$2.44$~h), of which $5 \times 32.15 = 161$~s is attributable to the agentic workflow, i.e., $1.8\%$ of the total wall-clock time, or a $\mathbf{1.9\%}$ overhead relative to the $8610$~s of EoH at an equal generation budget.

\begin{table}[h!]
\centering
\small
\begin{tabular}{llllll}
\toprule
\textbf{Block} & \textbf{Component} & \textbf{Avg.\ call (s)} & \textbf{\# calls} & \textbf{Total (s)} & \textbf{\% total} \\
\midrule
                & Reasoner (Hypothesis gen.) & 22.5 & 1 & 22.5 & 4.86 \\
Agentic         & Coder (Artifact generation) & 7.41 & 1 & 7.41 & 1.60 \\
Workflow        & Sandbox execution & 2.24 & 1 & 2.24 & 0.48 \\
                & \textit{Sub-total} & & & \textit{32.15} & \textit{6.95} \\
\midrule
Standard AHD   & Heuristic generation & 1.92 & 150 & 288.0 & 62.25 \\
steps          & Heuristic evaluation & 0.95 & 150 & 142.5 & 30.80 \\
               & \textit{Sub-total} & & & \textit{430.5} & \textit{93.05} \\
\midrule
\textbf{Total} & & & & \textbf{462.65} & \textbf{100} \\
\bottomrule
\end{tabular}
\caption{Average wall-clock breakdown for a single \name{} epoch in which the agentic workflow is executed ($I=150$ generation steps), over 10 independent runs. TRACE's Agentic Workflow incurs only 6.95\% of the total epoch duration; the remaining 93.05\% is attributed to the standard AHD steps shared with baselines such as EoH.}
\label{tab:cost_epoch}
\end{table}

\newpage
\section{LLM comparison}\label{app:llm_comparison}

This appendix assesses the generalizability of \name{} across LLM backbones and isolates the contribution of the proposed agentic architecture from that of any single model. To this end, we conduct two complementary, symmetric experiments: (i) varying the LLM backbones used inside \name{} (Sec.~\ref{sec:app_llm_trace}); and (ii) varying the LLM backbone of the most relevant AHD baseline, EoH (Sec.~\ref{sec:app_llm_eoh}), including the same frontier reasoning model used by \name{}'s Reasoner (\texttt{gemini-3-pro-preview}). All experiments use Scenario~1 with $3000$ generations over $10$ independent runs, matching the protocol of Sec.~\ref{sec:eval1}.

\subsection{LLM choice for \name{}}\label{sec:app_llm_trace}
\begin{figure}[h!]
\centering
\includegraphics[width=0.6\linewidth]{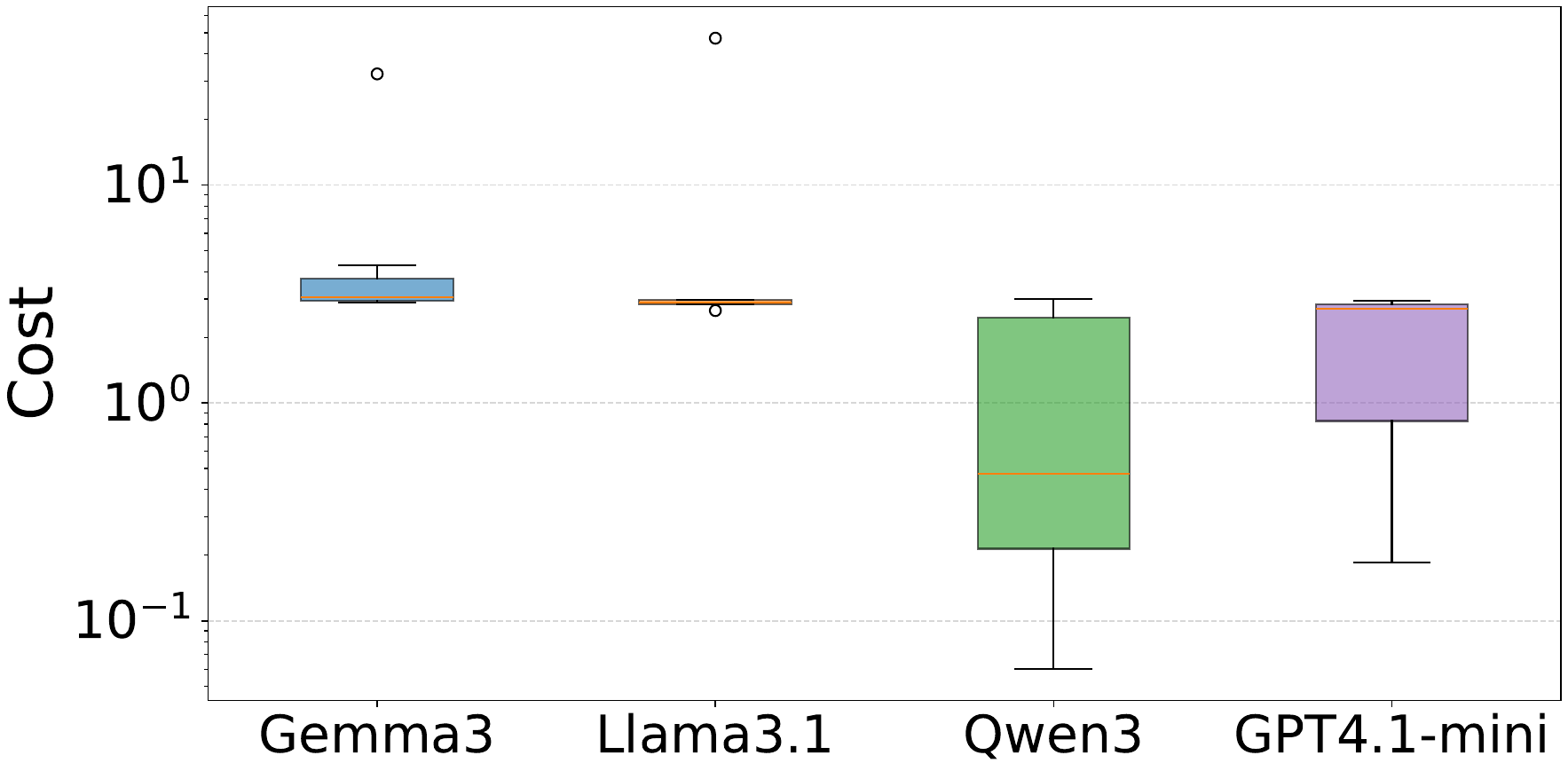}
\caption{LLM choice for \name{}'s Coder/Evolution loop: distribution of the best per-run cost across $10$ independent runs on Scenario~1.}\label{fig:llm_coder_distribution}
\end{figure}

We first evaluate \name{} across a spectrum of state-of-the-art open-weight and proprietary models used to instantiate the \textit{Coder Agent} and the \textit{heuristic-generation LLM}. Fig.~\ref{fig:llm_coder_distribution} reports the distribution of the best per-run cost across $10$ independent runs for four backbones: \texttt{gemma-3-27b-it}, \texttt{Llama-3.1-70B-Instruct}, \texttt{Qwen3-30B-A3B-Instruct}, and \texttt{GPT-4.1-mini-2024-12-01-preview}.

\texttt{Qwen3-30B} attains the lowest median cost (Eq.~\eqref{eq:cost}) and is therefore selected as the default backbone for the evaluations in Sec.~\ref{sec:eval1}--\ref{sec:5g}. The remaining backbones are competitive, with median costs only marginally higher.

%
The distribution also reveals two outliers, one each for \texttt{gemma-3-27b-it} and \texttt{Llama-3.1}, in which the final algorithm cost is substantially degraded. Inspection of the generation trace identifies a common root cause: a cascading failure in Diagnostic Artifact synthesis. In these runs, the model failed to produce syntactically correct Python code for the surrogate-model training pipeline within $M_\text{max}$ trials. Lacking a learned regressor, the heuristic-generation loop fell back to approximating processing latency from aggregate, text-based diagnostic feedback alone --- an approximation too coarse for the heterogeneous workload, leading to a sharp drop in efficiency.
This failure mode reinforces our central hypothesis: \name{}'s gains are contingent on the successful extraction and exploitation of system-log knowledge through the agentic workflow.

We now study the impact of the Reasoner backbone. While the main results use \texttt{gemini-3-pro-preview}, we re-ran the agentic workflow with \texttt{Claude-Opus-4.6} and \texttt{GPT-5.4}; Prompts~\ref{prompt:DH_claude}--\ref{prompt:DH_gpt} list the diagnostic hypotheses each model produced. Final \name{} performance is essentially unchanged across the three Reasoners. Inspection of the executed artifacts shows that only a small subset of hypotheses translates into measurable algorithmic gains, specifically: (i) training and querying a regression model that approximates processing time from task features and processor identity; (ii) ranking processors by aggregate performance (e.g., from fastest to slowest); and (iii) estimating per-processor queue length from drop events. Because all three Reasoners derive these three ideas independently, the resulting heuristics --- and therefore the final cost --- are comparable.

\subsection{LLM choice for the EoH baseline}\label{sec:app_llm_eoh}

We next evaluate the strongest AHD baseline under the same frontier reasoning model used inside \name{}'s Reasoner. We focus on EoH because, as noted in Sec.~\ref{sec:eval1}, \name{} without the agentic workflow is functionally equivalent to EoH; any residual gap between the two methods under matched backbones must therefore be attributable to the architectural component rather than to LLM capabilities.

\begin{figure}[h!]
\centering
\includegraphics[width=0.6\linewidth]{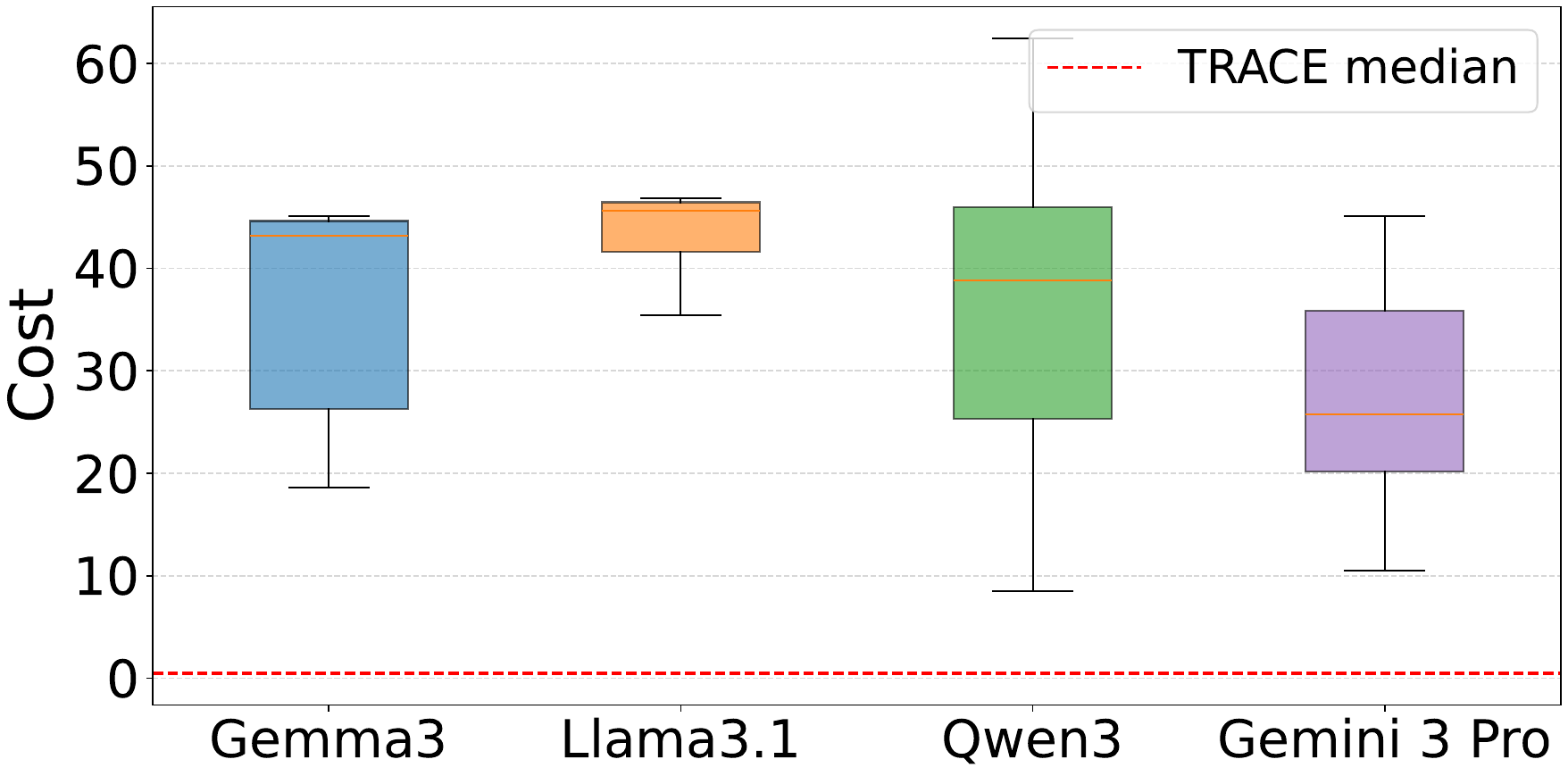}
\caption{EoH cost distribution across $10$ independent runs on Scenario~1 under four LLM backbones, including \texttt{gemini-3-pro-preview} (the model used by \name{}'s Reasoner). Even under the strongest backbone, EoH remains $1$--$2$ orders of magnitude above \name{} (cf.\ Fig.~\ref{fig:cost_dist}, left, where \name{}'s median cost is below~$1$).}\label{fig:eoh_llm_comp}
\end{figure}

Fig.~\ref{fig:eoh_llm_comp} reports the distribution of the best per-run cost across $10$ independent runs for \texttt{Gemma-3-27b-it}, \texttt{Llama-3.1-70B-Instruct}, \texttt{Qwen3-30B-A3B-Instruct}, and \texttt{gemini-3-pro-preview}.

Upgrading EoH's backbone from \texttt{Qwen3-30B} (median $\approx 38$, IQR $\approx [25, 46]$) to \texttt{gemini-3-pro-preview} (median $\approx 26$, IQR $\approx [20, 35]$) reduces median cost by roughly $30\%$ and tightens the run-to-run spread, but produces no qualitative jump. EoH with \texttt{gemini-3-pro-preview} remains roughly $30\times$ above \name{}'s median cost on the same scenario, and orders of magnitude above \name{}'s best-found solution reported in Table~\ref{tab:best}. The reason is that, even with a stronger backbone, EoH still operates exclusively on scalar fitness signals; it has no access to the system-dynamics information that \name{}'s agentic workflow extracts from the logs and injects into the heuristic-generation prompt. Combined with Sec.~\ref{sec:app_llm_trace}, where weakening \name{}'s Coder/Evolution backbone leaves performance essentially unchanged, the two controls converge on a single conclusion: the performance gap between \name{} and the AHD baselines is governed by the agentic architecture, not by the choice of LLM.

\begin{llmprompt}[label={prompt:DH_claude}]{Diagnostic Hypotheses generated using \texttt{Claude-Opus-4.6}.}
- Train a regression model per processor to predict proc_time from features.

- Cluster task features (e.g., KMeans) and identify each cluster's fastest processor from historical proc_time.

- Estimate each processor's queue capacity from the maximum observed queue\_states value before losses occur.

- Compute each processor's average proc_time to rank them by overall speed/capability.

- Estimate task arrival rate from time differences to anticipate congestion and preempt overloads.
\end{llmprompt}

\begin{llmprompt}[label={prompt:DH_gpt}]{Diagnostic Hypotheses generated using \texttt{GPT-5.4}.}
- Train a regression model predicting processing time per processor using task features.

- Discover feature-to-processor patterns that minimize execution time.

- Estimate processor-specific loss probability versus queue depth to infer safe load thresholds.

- Identify processors with consistently lower waiting times for similar feature clusters.

- Identify maximum observed queue depths per processor to estimate their individual storage capacity limits.
\end{llmprompt} 
\newpage
\section{Auditability Study}\label{app:audit}

This appendix details the LLM-as-judge auditability study referenced in Sec.~\ref{sec:audit}. We score the best-found heuristics from each method on three rubrics --- \textit{Semantic Grounding}, \textit{Auditability}, and \textit{Simulatability} --- using independent LLM judges.

\paragraph{Methodology.}
We employ a panel of three judges from distinct model families: \texttt{Gemini~3.1 Pro Preview}, \texttt{Claude~Opus~4.7}, and \texttt{GPT-5.5}. For each (judge, algorithm, rubric) triple we issue five independent queries at temperature $0.7$ and record the integer score $s \in \{1,\ldots,5\}$ returned by each reply; results are reported as the mean over those five queries. The study covers the best-found algorithms for Scenario~1 (Appendix~\ref{app:best_algo}).

\paragraph{Rubrics.}
Each rubric uses a 1--5 scale with the following definitions, applied to the same Python code submitted to the judge.

\begin{tcolorbox}[colback=white, colframe=black, boxrule=0.5pt, arc=2mm]
\textbf{Task 1: Semantic Grounding (Mechanism Identification).}\\
Analyze the scheduling algorithm. Evaluate the extent to which the logic and variables are grounded in physical system states versus abstract, system-agnostic mathematical transformations.
\begin{itemize}
\item[\textbf{1}] Completely abstract. Relies entirely on ``magic numbers'' or arbitrary mathematical transformations.
\item[\textbf{2}] Mostly abstract, but uses standard normalization or minor structural logic.
\item[\textbf{3}] Mixed. Uses some domain-relevant logic, but heavily relies on opaque weights or mathematical scrambling.
\item[\textbf{4}] Mostly grounded. The logic clearly attempts to model system dynamics, with only minor arbitrary scaling factors.
\item[\textbf{5}] Fully grounded. Every step of the algorithm maps directly to a clear, physical system concept.
\end{itemize}
\end{tcolorbox}

\begin{tcolorbox}[colback=white, colframe=black, boxrule=0.5pt, arc=2mm]
\textbf{Task 2: Auditability (Locating Safety-Relevant Logic).}\\
A critical safety requirement is that no single processor should be overloaded or saturated. Without executing the code, estimate how quickly a competent operator could (a) locate the lines that govern processor saturation and (b) identify the exact edit required to tighten or loosen the saturation threshold of a chosen processor. Score on operator effort, not on the syntactic form of the code.
\begin{itemize}
\item[\textbf{1}] Operator cannot locate any saturation-governing logic; load distribution is emergent rather than encoded.
\item[\textbf{2}] Operator can identify a generic load-distribution mechanism (e.g., round-robin) but cannot point to any per-processor threshold.
\item[\textbf{3}] Operator can locate saturation-relevant logic, but it is entangled with unrelated computation; isolating the change site requires careful reading.
\item[\textbf{4}] Operator can locate the relevant lines quickly; the change site is identifiable but the threshold is parameterised through indirect computation.
\item[\textbf{5}] Operator can locate, in under 30 seconds of reading, the exact line(s) whose modification would change the saturation threshold of a specific processor.
\end{itemize}
\end{tcolorbox}

\begin{tcolorbox}[colback=white, colframe=black, boxrule=0.5pt, arc=2mm]
\textbf{Task 3: Simulatability (Trace Prediction).}\\
You are given a fixed scenario: a feature vector $\mathbf{x}$ with all entries equal to $1.0$, a queue-occupancy vector $\mathbf{q}=\mathbf{0}$, and the processor count specified by the algorithm. Without executing the code, (i) predict the processor index the algorithm assigns the task to and (ii) report one numeric intermediate (e.g., a per-processor score) that any other reader of the same code would also compute. Score the algorithm on how reliably an \emph{external} engineer, given only the code and this scenario, would arrive at the same prediction.
\begin{itemize}
\item[\textbf{1}] No external engineer could reproduce the prediction without running the code; the output depends on opaque non-linear transformations of the input.
\item[\textbf{2}] An engineer could guess the rough output range, but not the assigned processor.
\item[\textbf{3}] An engineer could narrow the assigned processor to a small subset; exact selection depends on hard-to-trace interactions.
\item[\textbf{4}] An engineer could reproduce the assigned processor, with minor uncertainty in tie-breaking or numeric rounding.
\item[\textbf{5}] Two independent engineers reading the code would deterministically arrive at the same processor index and the same intermediate score.
\end{itemize}
\end{tcolorbox}

\paragraph{Results.}
Table~\ref{tab:audit} reports the per-judge scores summarised in
Sec.~\ref{sec:audit}.
\name{} is the only method that scores above 1.3 on Semantic Grounding: all three judges assign~3 or~4, recognising that the algorithm's branching logic and penalty factors are tied to predicted processing times and a median-relative threshold, while every baseline converges to abstract transformations (1/5 from almost all judges; the only exception is HSEvo, which receives 1.80 from GPT). On Auditability, \name{} scores 4--5 across the panel, reflecting the ability of an operator to easily locate and modify the important parts of the algorithm. Baselines score~1--2; even ReEvo, which distributes load via a weighted dot-product, offers no per-processor threshold to change.

Regarding Simulatability, \name{} scores 1/5 unanimously. This is expected: the discovered algorithm calls to \texttt{predict()}, which makes mental trace-through impossible without execution. Conversely, ReEvo achieves a unanimous 5/5: its three-line modulo policy (\texttt{int(abs(dot(features,[1,2,3,4,5]))\%10)}) can be evaluated by hand in seconds for any fixed input. EoH also achieves a high score for the same reason: once the fixed \texttt{np.random.seed(42)} coefficients are materialised, the computation is a single \texttt{tanh} followed by a scale-and-floor. The Simulatability column therefore does not reflect auditability in the operational sense: it captures how easily execution can be mentally simulated, which correlates with structural simplicity, not with the ability to reason about or modify the algorithm's safety behaviour.

\begin{table}[h!]
\centering
\resizebox{\textwidth}{!}{%
\begin{tabular}{lcccccccccc}
\toprule
Strategy
  & \multicolumn{3}{c}{Semantic Grounding}
  & \multicolumn{3}{c}{Auditability}
  & \multicolumn{3}{c}{Simulatability}
  & Mean \\
\cmidrule(lr){2-4}\cmidrule(lr){5-7}\cmidrule(lr){8-10}
  & Gemini & Claude & GPT
  & Gemini & Claude & GPT
  & Gemini & Claude & GPT
  & \\
\midrule
FunSearch
  & 1.00 & 1.00 & 1.00
  & 1.40 & 1.00 & 2.00
  & 1.00 & 4.00 & 1.00
  & \textbf{1.49} \\
EoH
  & 1.00 & 1.00 & 1.00
  & 1.00 & 1.00 & 2.00
  & 3.40 & \textbf{5.00} & \textbf{5.00}
  & \textbf{2.27} \\
ReEvo
  & 1.00 & 1.00 & 1.00
  & 2.00 & 2.00 & 2.00
  & \textbf{5.00} & \textbf{5.00} & \textbf{5.00}
  & \textbf{2.67} \\
HSEvo
  & 1.00 & 1.00 & 1.80
  & 1.00 & 1.00 & 2.00
  & 1.00 & 3.80 & 2.00
  & \textbf{1.62} \\
\textbf{\name{}}
  & \textbf{4.00} & \textbf{3.00} & \textbf{3.00}
  & \textbf{5.00} & \textbf{4.00} & \textbf{5.00}
  & 1.00 & 1.00 & 1.00
  & \textbf{3.00} \\
\bottomrule
\end{tabular}%
}
\caption{Multi-judge auditability scores for the best-found heuristics in Scenario~1.
Each cell reports the mean over five queries at temperature~$0.7$;
the Mean column averages the nine (rubric, judge) means per method.}
\label{tab:audit}
\end{table}

\newpage
\section{Real-World Case Study: Energy-Efficient 5G Virtualization}
\label{app:mobile_motivation}

\subsection{Motivation and Problem Formulation}

To demonstrate the practical utility of \name{} in industrial-grade environments, we instantiate the general resource assignment framework within the domain of 5G virtualized radio access networks (vRAN). This setup grounds our analysis on a real-world use case where the system must consistently satisfy hard time deadlines for signal processing while minimizing the energy toll.

In modern 5G architectures, base station (BS) functions are decoupled from proprietary hardware and executed as software processes on general-purpose cloud computing platforms at the edge~\citep{cloudric}; see Fig.~\ref{fig:5g_usecase}. The computational bottleneck of this pipeline resides in the Physical Layer (PHY) processing---specifically the decoding of uplink user data encapsulated in Transport Blocks (TBs). Upon arrival, raw radio signals must undergo Forward Error Correction (FEC), typically utilizing LDPC decoding algorithms, to recover the original information.

This process is highly computationally intensive and subject to hard time deadlines, typically in the 1--3~ms range; failing to decode a TB within this budget results in data loss and compromises industry-grade reliability. Moreover, as 5G networks densify, the energy consumption of these decoding tasks has emerged as a primary operational expenditure (OPEX) and environmental concern for operators. Consequently, it is of paramount importance to \emph{schedule} appropriate computing resources to 5G wireless processing tasks to minimize the energy footprint while guaranteeing carrier-grade processing deadlines.

We model the stream of incoming TBs as a task sequence. Each task $i$ is characterized by a feature vector $\mathbf{x}_i \in \mathbb{R}^3$, comprising three physical layer parameters that dictate computational complexity:
\begin{itemize}
    \item \textit{Transport Block Size (TBS)}: The data volume represented by the total number of bits. Larger TBs typically lead to increased processing latency as the decoding workload scales with data volume.
    \item \textit{Modulation and Coding Scheme (MCS)}: The spectral efficiency index, indicating the density of the signal constellation. Higher MCS levels are more spectral efficient (carry more bits per signal sample) but may be more compute-intensive depending on the signal quality.
    \item \textit{Signal-to-Noise Ratio (SNR)}: The quality of the received signal, measured in dBs, that depends on the radio channel. Lower SNR necessitates more iterative convergence steps in the decoding algorithm, non-linearly increasing processing time.
\end{itemize}

An edge computing platform may comprise heterogeneous processing units, including general-purpose CPUs and specialized hardware accelerators (HAs) such as GPUs, ASICs, or FPGAs. To characterize the hardware heterogeneity inherent in this domain, we analyze the experimental data gathered from a production-grade vRAN testbed~\citep{cloudric}. The testbed consists of a pool of 16$\times$ Intel Xeon Gold 6240R CPU cores running the Intel FlexRAN library to process signal processing requests, and an NVIDIA V100 GPU utilizing a commercial LDPC driver (NVIDIA Aerial) to accelerate decoding tasks.

\begin{figure}[h!]
    \centering
    \includegraphics[width=0.85\linewidth]{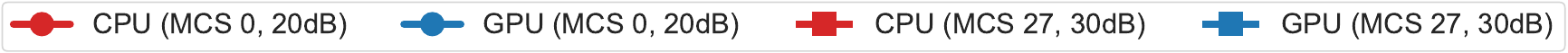} 
    
    \begin{subfigure}[b]{0.49\linewidth}
        \centering
        \includegraphics[width=0.8\linewidth]{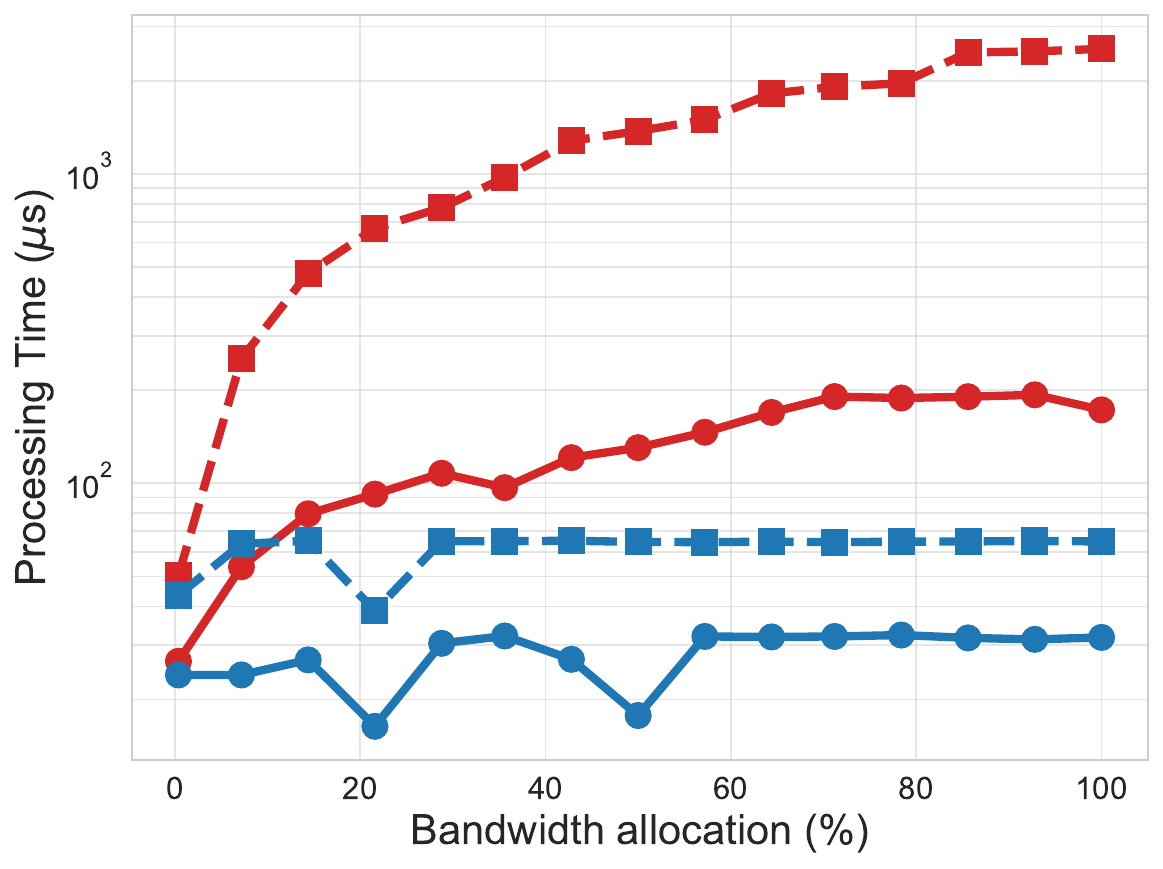} 
        \caption{}
        \label{fig:5g_time}
    \end{subfigure}
    \begin{subfigure}[b]{0.49\linewidth}
        \centering
        \includegraphics[width=0.8\linewidth]{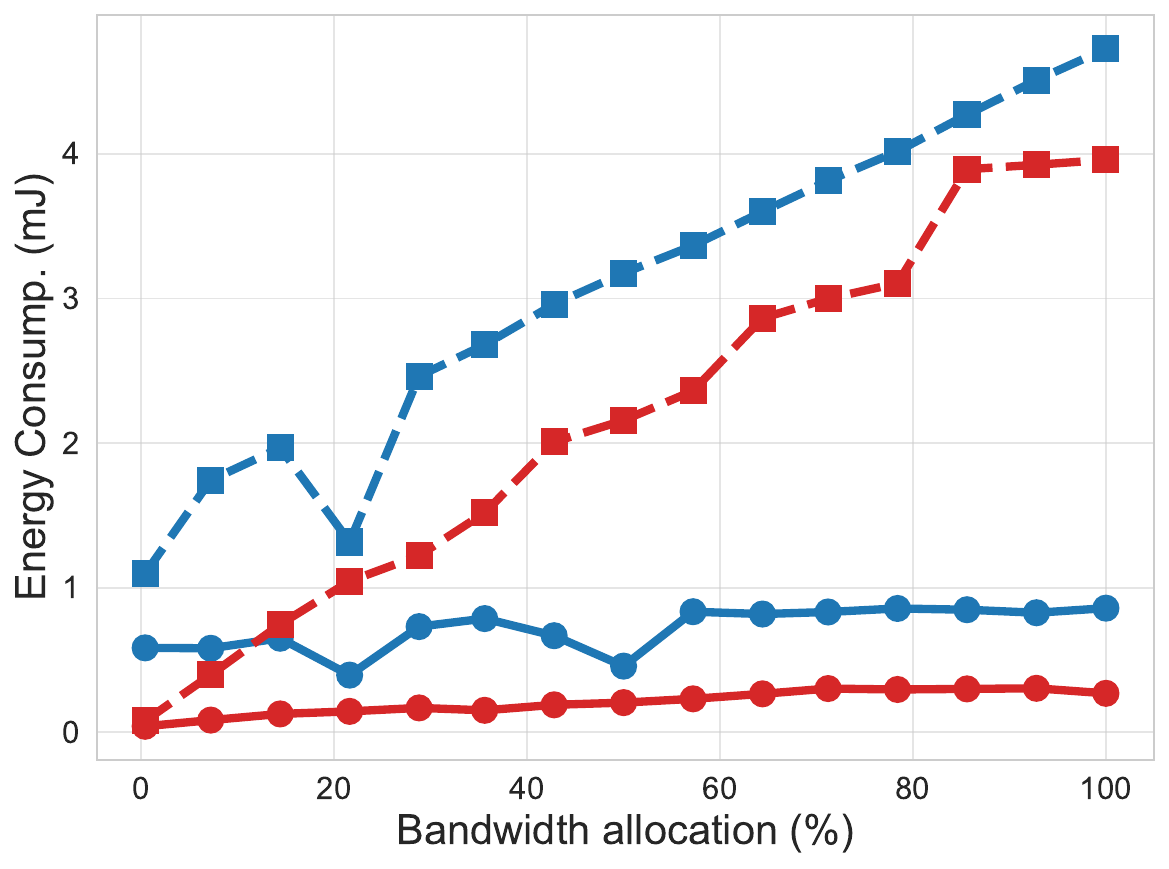}
        \caption{}
        \label{fig:5g_energy}
    \end{subfigure}
    \caption{Latency-energy trade-off in heterogeneous 5G edge clouds.}
    \label{fig:5g_motivation}
\end{figure}

Fig.~\ref{fig:5g_motivation}(a) illustrates the latency-energy trade-offs: CPU processing latency scales linearly with the allocation of bandwidth resource (and so larger TBs) frequently exceeding hard time deadlines for large workloads. In contrast, the GPU leverages massive parallelism to maintain near-constant processing times, effectively absorbing high-demand workloads without saturation.
However, this performance advantage incurs a significant energy toll. As shown in Fig.~\ref{fig:5g_motivation}(b), the CPU maintains significantly higher energy efficiency compared to the GPU-based HA, particularly for smaller TBs. We further note that these dynamics are governed by the Signal-to-Noise Ratio (SNR). A lower SNR necessitates additional iterative convergence steps in the belief-propagation decoding algorithm implemented by LDPC decodders, non-linearly increasing both processing latency and energy consumption.

To address this use case, we adapt the general problem formulation from Sec.~\ref{sec:prob_form} to reflect the system constraints of a 5G vRAN. In this context, the optimization objective shifts from raw latency minimization to \textit{energy efficiency}, subject to a hard processing delay constraint. Latency is treated as a binary feasibility condition rather than a continuous cost. Accordingly, we redefine the cost $C_i$ for a task $i$ assigned to processor $n$ as:
\begin{equation} \label{eq:cost_mobile}
    C_i = \begin{cases} 
    r_{i,n} & \text{if } t_{i,n} + t^{\text{wait}}_{i,n} \leq D \\
    \Gamma & \text{otherwise (Deadline Violation)}
    \end{cases}
\end{equation}
where $r_{i,n}$ is the energy consumption of computing node $n$ when processing task $i$, $t_{i,n}$ is the processing time of task $i$ in processor $n$, $t^{\text{wait}}_{i,n}$ is the waiting time of task $i$ in processor $n$'s queue, $D$ is the time deadline imposed by system requirements, and $\Gamma$ is a large penalty representing the service violation cost.

\subsection{Experimental Setup}

\begin{figure}[ht!]
    \centering
        \includegraphics[width=0.75\linewidth]{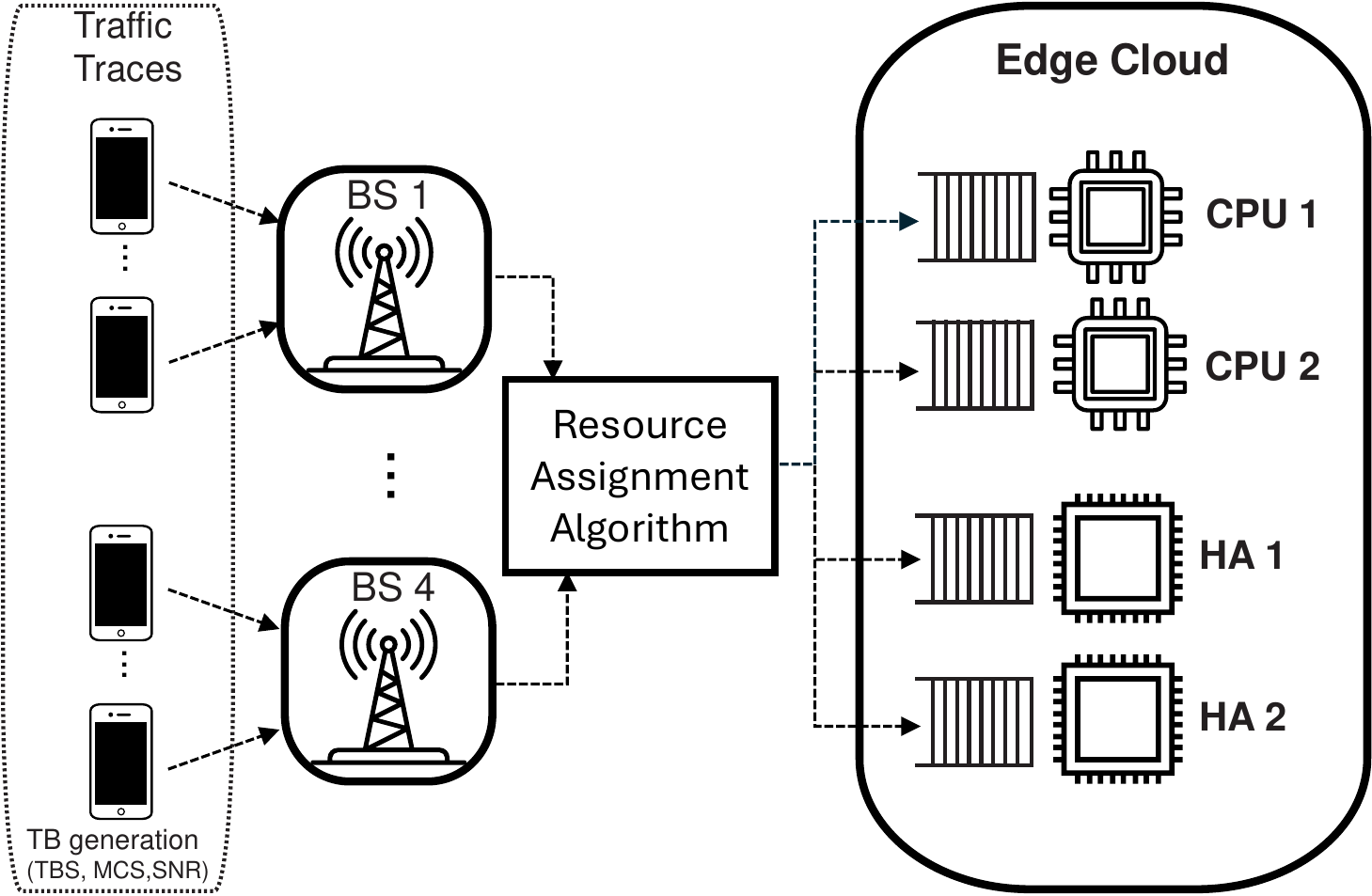}
    \caption{Energy-Efficient 5G Scheduling}
        \label{fig:5g_usecase}
\end{figure}

To evaluate the framework under realistic conditions, we simulate a heterogeneous edge computing cluster consisting of four processing units, denoted as $\mathcal{N} = \{\text{CPU}_1, \text{CPU}_2, \text{HA}_1, \text{HA}_2\}$, and illustrated in Fig.~\ref{fig:5g_usecase}. The set comprises two  CPU processors and two hardware accelerators (HAs) based on the cloud system analyzed earlier. To capture the manufacturing variability and generation upgrades typical in data centers, we introduce both inter-class and intra-class heterogeneity:
\begin{itemize}
    \item \textit{Inter-class:} CPUs and HAs exhibit distinct energy-latency profiles as detailed in Fig.~\ref{fig:5g_motivation}. While CPUs offer superior energy efficiency for smaller workloads, HAs provide higher performance, required to meet stringent deadlines for large TBs.
    \item \textit{Intra-class:} The accelerator pool is asymmetric; $\text{HA}_1$ represents a next-generation unit, modeled with a 20\% improvement in both processing throughput and energy efficiency compared to $\text{HA}_2$.
\end{itemize}
To ensure the simulation fidelity extends beyond theoretical models, the processing-time and energy distributions $T^{(n)}$ and $R^{(n)}$ are replaced by lookup queries to a real-world dataset~\citep{cloudric_dataset}. This dataset contains empirical measurements from an industrial-grade vRAN platform, preserving the complex non-linearities.

The input traffic emulates a dense multi-cell environment comprising four BSs, each operating with a maximum 100-MHz bandwidth and serving four active users. We adopt a \textit{trace-driven approach} to generate the request stream. Rather than synthesizing traffic from static distributions, we replay anonymized user traces collected from an operational mobile network using open-source monitoring tools~\citep{cloudric}. 
These traces dictate the temporal arrival of signal processing requests (TBs) and their associated properties, specifically TBS, SNR, and MCS. This methodology ensures that our evaluation captures the temporal correlations, sub-millisecond burstiness, and channel fading dynamics inherent to real-world mobile environments.

The system interface is aligned with the Open RAN (O-RAN) Alliance specifications\footnote{\url{https://www.o-ran.org/specifications}}, where the scheduling logic has access to the instantaneous queue occupancy across all heterogeneous processing units. However, the exact mapping between task features $\mathbf{x}_i$ and the resulting processing latency or energy expenditure remains hidden and must be inferred online. The system operates under a hard real-time constraint typical of 5G carrier-grade deployments, with a strict processing deadline of 1~ms; any Transport Block (TB) that fails to complete decoding within this TTI-level window is discarded, incurring a reliability violation. To formalize the trade-off between energy efficiency and constraint satisfaction, we set the penalty multiplier to $\Gamma=10$ for the cost computation, effectively prioritizing the reliability target.

\newpage
\section{Prompts}\label{app:prompts}

\subsection{Definitions}

At each generation step $i$, \name{} synthesizes a \textit{Generation Prompt} to guide the LLM in producing a new candidate algorithm. This prompt acts as a structured context window, integrating static problem definitions with dynamic environmental insights. It comprises four essential components:

\textbf{1. Problem Description:} A formal natural language definition of the resource allocation environment, encompassing invariant system parameters (e.g., computing node count $N$, feature dimensionality $d$) and the input/output interface required for the algorithm.
    
\textbf{2. System Information:} A dynamic textual description of the environment's latent properties, updated at the start of each epoch. This component encapsulates the \textit{Diagnostic Knowledge} extracted by the agentic workflow, effectively defining the boundary between known and unknown system dynamics. It includes derived insights (e.g., "Processor 0 is more efficient with a certain type of tasks") and inferred constraints, providing the LLM with a grounded inductive bias for optimization.
    
\textbf{3. Example Algorithms:} A set of $m$ parent heuristics sampled from the database. To drive improvement, selection is biased towards high-scoring candidates.
    
\textbf{4. Evolutionary Operators:} Meta-instructions that define the variation strategy for the current step. Adopting methodologies from recent AHD literature~\citep{eoh, reevo}, these operators instruct the LLM to perform crossover (recombining logic from parents) or mutation (introducing novel code structures), thereby balancing exploration and exploitation.

\subsection{Prompt examples}

\begin{llmprompt}[label={prompt:prob_descr}]{Problem description prompt. Scenario 1 of the general task scheduling problem.}
You are tasked with designing an algorithm that maps an incoming task to a specific processing unit.
Input:
- A task characterized by a `features` vector (NumPy array of size 5).
- 4 processing units, numbered 0 to 3.

System Dynamics:
1. Heterogeneity: Each processor has different hardware capabilities. A specific task's features combined with a specific processor determine the Processing time (service time). 
2. Queueing: Tasks arrive sequentially. If a processor is busy, the task enters its queue. 
3. Loss Mechanism: A task is only lost if the assigned processor's queue is already full at the moment of arrival. Tasks are NOT lost based on their features.

Objective:
Your solution must optimize for the following goals:
1. Minimize the total number of lost tasks (prevent queue overflows).
2. If losses are minimized (or zero), minimize the Total Time in System for each task. 
Total Time = Waiting Time (in queue) + Processing Time (execution). 
\end{llmprompt}

\begin{llmprompt}[label={prompt:system_info}]{System Information prompt. Scenario 1 of the general task scheduling problem.}
System information:
- The exact relationship between task features and processing speed is unknown.
- The queue lengths (how many elements can be stored in a queue) are unknown. 
- The current queue depths (how many elements are in the queue) are unknown.
\end{llmprompt}


\begin{llmprompt}[label={prompt:system_coder}]{System instructions prompt for Coder agent.}
You are an AI that responds with only Python code. At the end of your response, provide the Python code in a code block. For example: ```python\nprint('Hello world!')```

Implement a Python function called select\_pu. This function accepts one input: `features`, which is a numpy array. The function returns one integer that indicates the selected processing unit.
\end{llmprompt}


\begin{llmprompt}[label={prompt:system_reasoner}]{System instruction prompt for Reasoner agent.}
You are an experienced, data-driven system engineer. You will be asked to provide useful ideas and insights based only on the provided data to help in the design of algorithms.
\end{llmprompt}


\begin{llmprompt}[label={prompt:DH_generation}]{Diagnostic Hypothesis generation prompt for Reasoner agent.}
[Problem description prompt]

[System information prompt]

To solve this problem, you can leverage the knowledge from the logs of the system to optimize.
The log is a CSV file with 5 columns separated by comma `,`. Each row contains the information of one task processed in the system with the following information: 
- time: the arriving time of the task to the system.
- features: a vector with the features of the task.
- lost: 0 if the task is successfully processed or 1 if the task is lost.
- proc_time: processing time of the task.
- proc: index of the processing unit that processed the task.
- queue_states: number of tasks in the queue of each processor when the task arrives.

[So far, you have proposed the following ideas to exploit the log data:
    (List of previous Diagnostic hypotheses)
Now, you need to propose a new idea different from the previous ones.] # in case there are previous Diagnostic hypotheses

[Based on the log's data, propose a simple idea (under 15 words) to provide an insight from the data that can be useful for decision making.] # for semantic insight

[Based on the log's data, propose a simple idea (under 15 words) to train a simple model (e.g., from sklearn) that can be used as a component for decision making.]  # for surrogate models
\end{llmprompt}


\begin{llmprompt}[label={prompt:DA_1}]{Diagnostic Artifact generation prompt (first part) for Coder agent.}
[Problem description prompt]

[System information prompt] 

Write the Python code to implement the following task:
(Diagnostic hypothesis)

Use as input the log file located at: ```/route/to/log/log_XXX.csv'''
The format of the log file is:
time,features,lost,proc_time,proc,queue_states
7.0,[0.84312234 0.18156817 0.17044879 0.6092944  0.70704929],0,8.307854811495774,2,"[1 1 0 0 0 0 0 0 0 0 0 1 2 1 0 2 2 0 0 1 1 0 0 1 1 0 0 0 1 0 0 0 0 0 0 0 0 0 0 0 0 1 0 0 0 0 1 0 0 0]"
22,[0.85748415 0.24441935 0.15581696 0.19658431 0.80626611],0,4.531390180952061,3,"[2 1 0 0 2 0 0 0 1 0 0 0 5 2 0 3 2 0 0 0 0 0 1 0 1 1 2 0 1 0 0 0 0 0 0 1 2 0 1 0 1 2 0 0 0 0 1 1 0 0]"
28,[0.33265489 0.07972651 0.62192413 0.43560503 0.17603169],0,1.0869485908215089,3,"[1 1 0 0 2 0 1 1 1 0 1 0 2 2 1 4 2 0 0 0 1 0 1 0 0 1 2 0 0 0 0 0 0 1 0 2 2 0 2 0 2 2 0 0 0 0 1 1 0 0]"
\end{llmprompt}


\begin{llmprompt}[label={prompt:DA_2_surrogate}]{Diagnostic artifact generation prompt (second part for surrogate model) for Coder agent.}
- Step 1: Generate the code to load the data, train the model, and save it on disk in the folder: /path/to/model/weights/model_XXXX/.
- Step 2: Write a function to use the model, loading the model from disk and adding all the imports needed.
- Step 3: Provide an example of usage of the previous function.

Example:

```python
# Step 1
import pandas as pd
import numpy as np
from sklearn import model

# Load the data
log_file =/path/to/log.csv'
df = pd.read_csv(log_file)
...

# Train the model
model = Model()
model.fit(X, y)

# Save the model
model.save(...)

# Step 2
import numpy as np
# add all the needed imports
model_weights = ...

def predict(new_task_features, proc_id):
    model = model.load(model_weights)
    ...
    return prediction

# Step 3
prediction = predict(new_task_features, proc_id)
\end{llmprompt}


\begin{llmprompt}[label={prompt:DA_3_insight}]{Diagnostic artifact generation prompt (second part for semantic insight) for Coder agent.}
- Step 1: Generate the code to load and process the log data.
- Step 2: Use the Python function `print()' to output insights from the system extracted from the processed data.


Example:

[Step 1]
```python
import pandas as pd
import numpy as np
from sklearn.linear_model import LinearRegression

# Load the data
log_file =/path/to/log.csv'
df = pd.read_csv(log_file)
...
```

[Step 2]
```python
...
for i in range(n_proc):
 print(f"Queue length of processor {{i}}: {{queue_len[i]}}")
```
\end{llmprompt}

\begin{llmprompt}[label={prompt:log_example}]{Excerpt from a CSV log file obtained from Scenario 1 execution.}
time,features,lost,proc_time,proc,queue_states
10.000000000000002,[0.37340462 1.         0.         0.         0.        ],0,10.186702309090437,9,[0 3 0 2 3 1 0 1 6 2]
34.00000000000002,[0.58789216 1.         0.         1.         0.        ],1,50.73486520211199,5,[ 2  5  4 10  7 10  1  6  9  8]
34.00000000000002,[0.55916644 1.         0.         0.         0.        ],1,25.69895804511365,5,[ 2  5  4 10  7 10  1  6  9  8]
31.00000000000002,[0.94553469 0.         1.         0.         0.        ],0,3.363836737190401,0,[ 1  5  3  7  6 10  2  5 10  8]
36.00000000000003,[0.59170117 0.         0.         0.         1.        ],1,10.295850586337183,8,[ 1  5  4 10  7 10  2  6 10  9]
36.00000000000003,[0.30450687 0.         0.         0.         1.        ],1,40.22534354893957,1,[ 1  5  4 10  7 10  2  6 10  9]
37.00000000000003,[0.86630185 0.         0.         1.         0.        ],1,26.08287731226671,5,[ 1  5  4 10  8 10  3  6 10  9]
37.00000000000003,[0.01060544 1.         0.         1.         0.        ],1,50.53027220661059,1,[ 1  5  4 10  8 10  4  6 10  9]
37.00000000000003,[0.21383168 0.         1.         0.         1.        ],1,25.292289597469328,5,[ 1  5  4 10  8 10  5  6 10  9]
33.00000000000002,[0.22715506 0.         1.         0.         1.        ],0,2.567887659375335,0,[ 1  5  4  9  7 10  1  6  9  7]
38.00000000000003,[0.83845124 0.         0.         0.         1.        ],1,10.419225622097883,8,[ 0  5  4 10  8 10  5  6 10  9]
38.00000000000003,[0.4205742  0.87895317 0.63688969 0.95158695 0.8096467 ],1,24.20031002335317,8,[ 0  5  4 11  8 10  6  6 10  9]
\end{llmprompt}

\begin{llmprompt}[label={prompt:log_example_5G}]{Excerpt from a CSV log file obtained from 5G virtualization use case execution.}
time,features,lost,proc_time,proc,queue_states,energy
188.00000000000014,[24. 27.  3.],0,0.0796905,0,[1 1 6 4],125.17559982774308
188.00000000000014,[24. 27.  3.],0,0.04096000000000001,2,[1 1 3 2],829.2038206303748
188.00000000000014,[24. 27.  3.],0,0.051199999999999996,3,[1 1 6 3],1295.6309697349604
188.00000000000014,[24. 27.  3.],0,0.04096000000000001,2,[1 1 4 2],829.2038206303748
188.00000000000014,[24. 27.  3.],0,0.0796905,1,[3 2 6 4],125.17559982774308
188.00000000000014,[24. 27.  3.],0,0.0796905,0,[2 1 6 4],125.17559982774308
188.00000000000014,[24. 27.  3.],0,0.04096000000000001,2,[1 1 5 2],829.2038206303748
216.00000000000017,[23. 28.  6.],0,0.0495616,2,[0 0 0 0],1003.3366229627534
216.00000000000017,[23. 28.  6.],0,0.06195199999999999,3,[0 1 1 0],1567.7134733793018
216.00000000000017,[23. 28.  6.],0,0.0495616,2,[0 1 1 2],1003.3366229627534
\end{llmprompt}

\newpage

\section{Best generated algorithms} \label{app:best_algo}
This Appendix includes the generated algorithms with the best performance for each strategy discussed in Sec.~\ref{sec:cloud_computing}.


\begin{pyalgorithm}[label={alg:ours_scenario_1}]{Best algorithm found by \name{}.}
def select_pu(features):
    # Core idea: Balance speed and stability by selecting processors that are fast but not excessively so,
    # while favoring those with higher robustness (lower variance in processing time and better feature alignment).
    # The strategy avoids overloading the fastest processor (P0) when it's significantly faster than median,
    # and instead promotes processors that offer a good trade-off between speed, reliability, and stability.

    import numpy as np
    np.random.seed(42)  # For reproducibility

    # Predict processing times for all 10 processors
    predicted_times = [predict(features, proc_id) for proc_id in range(10)]
    median_time = np.median(predicted_times)

    # Identify the fastest processor and assess if it's too fast (risk of overloading)
    min_time = min(predicted_times)
    best_pu = predicted_times.index(min_time)

    # If the fastest processor (P0) is much faster than median, avoid overloading it
    if best_pu == 0 and min_time < median_time * 0.5:
        # Exclude P0 and pick the next best among the rest
        candidates = [i for i in range(10) if i != 0]
        candidate_times = [predicted_times[i] for i in candidates]
        best_pu = candidates[np.argmin(candidate_times)]

    # If not P0, but still significantly faster than median, apply mild penalty
    elif min_time < median_time * 0.7 and best_pu != 0:
        # Apply a small penalty to very fast processors to discourage overuse
        # This helps prevent queue overflow by distributing load
        penalty_factor = 1.0 + 0.3 * (1.0 - min_time / median_time)
        # Re-evaluate with penalty
        adjusted_times = [t * penalty_factor if i == best_pu else t for i, t in enumerate(predicted_times)]
        best_pu = np.argmin(adjusted_times)

    # Add jitter for tie-breaking
    jitter = np.random.uniform(0, 0.0003)
    effective_cost = predicted_times[best_pu] + jitter
    # Final check: ensure we don't pick a processor with very high variance (unstable)
    # Use synthetic variation to estimate variance
    synthetic_features = np.random.randn(100, 5)
    variances = []
    for proc_id in range(10):
        times = [predict(synth_feat, proc_id) for synth_feat in synthetic_features]
        variances.append(np.var(times))
    
    # Normalize variance to reliability
    max_variance = max(variances) if max(variances) > 0 else 1
    reliability_scores = [1.0 - (v / max_variance) for v in variances]
    reliability_scores = [r / max(reliability_scores) for r in reliability_scores]
    # If the selected processor has low reliability, consider a more stable alternative
    if reliability_scores[best_pu] < 0.4:
        # Find the most reliable processor among the top 3 fastest
        top_k = 3
        sorted_indices = np.argsort(predicted_times)[:top_k]
        candidates = [i for i in sorted_indices if reliability_scores[i] >= 0.5]
        if candidates:
            best_pu = np.random.choice(candidates)  # Random choice among reliable fast ones
    return best_pu
\end{pyalgorithm}

\begin{pyalgorithm}[label={alg:best_funsearch}]{Best algorithm found by Funsearch.}
def select_pu(features):
    import numpy as np
    
    # Advanced deterministic hashing with maximum entropy, load balancing, and robustness
    # Designed for stateless, memoryless operation under complete uncertainty
    
    # Step 1: Apply diverse, non-linear, and high-sensitivity transformations
    # Each feature is transformed using a different mathematical function
    # to maximize entropy and prevent clustering under unknown system dynamics
    # Use small noise for symmetry breaking (deterministic via fixed seed)
    np.random.seed(42)
    noise = np.random.uniform(-1e-8, 1e-8, size=features.shape)
    x = features + noise
    
    # Apply a mix of non-linear functions with different scales and sensitivities
    transformed = np.array([
        np.sin(x[0]) * 1e15,                       # High-frequency oscillation
        np.cos(x[1]) * 1e14,                       # Large amplitude variation
        np.tanh(x[2]) * 1e13,                      # Extreme saturation for stability
        np.exp(np.clip(x[3], -15, 15)) * 1e12,     # Controlled exponential amplification
        np.log1p(np.abs(x[4]) + 1e-10) * 1e16      # Logarithmic scaling with epsilon
    ])
    
    # Step 2: Use a set of large, distinct, co-prime prime numbers
    # These primes are carefully selected to ensure maximal mixing and avoid periodicity
    primes = np.array([10007, 10009, 10037, 10039, 10061])
    
    # Step 3: Compute a weighted sum with prime coefficients
    # This creates a highly sensitive function of the input features
    weighted_sum = np.dot(transformed, primes)
    
    # Step 4: Apply chaotic-like mixing with large prime modulus and non-linear offset
    # Use 1000000007 (a large prime) and a large, non-repeating offset
    # Add a tiny jitter to prevent zero or symmetric cases
    hash_val = (weighted_sum * 1000003 + 987654321 + 1e-10) % 1000000007
    
    # Step 5: Final normalization and selection
    # Ensure non-negative result and map to 0-9 using modulo
    return int(abs(hash_val) % 10)
\end{pyalgorithm}

\begin{pyalgorithm}[label={alg:best_eoh}]{Best algorithm found by EoH.}
def select_pu(features):
    import numpy as np
    
    # Use the raw features to compute a deterministic hash-like value
    # This avoids reliance on normalization or variance-based weighting
    # Instead, we use a fixed set of random coefficients to create a unique fingerprint
    
    # Fixed seed for reproducibility in the hashing process
    np.random.seed(42)
    
    # Generate a set of random coefficients based on the feature length
    # This ensures the mapping is deterministic but complex
    coefficients = np.random.rand(5) * 2 - 1  # Random weights in [-1, 1]
    
    # Compute a weighted sum of features with the random coefficients
    hash_value = np.dot(features, coefficients)
    
    # Apply a non-linear transformation to spread the values
    # Use tanh for bounded output and strong non-linearity
    transformed = np.tanh(hash_value)
    
    # Scale to [0, 10) and take floor to get a PU ID
    pu_id = int((transformed + 1) * 5)  # maps [-1,1] to [0,10)
    
    # Ensure result is within [0, 9] using modulo
    return pu_id % 10
\end{pyalgorithm}

\begin{pyalgorithm}[label={alg:best_reevo}]{Best algorithm found by ReEvo.}
def select_pu(features):
    # Use a weighted sum of features with a fixed seed to introduce variance
    # This avoids deterministic bias while maintaining consistency
    # The weighting emphasizes feature variance and ensures robust load balancing
    # under unknown queue dynamics and system heterogeneity
    np.random.seed(42)  # Fixed seed for deterministic behavior despite randomness
    weights = np.array([1, 2, 3, 4, 5])  # Emphasize higher variance in later features
    weighted_sum = np.dot(features, weights)
    return int(np.abs(weighted_sum) % 10)
\end{pyalgorithm}

\begin{pyalgorithm}[label={alg:best_HSEvo}]{Best algorithm found by HSEvo.}
def select_pu(features):
    # Derive deterministic seed from multi-statistic input fingerprint
    seed = int((np.sum(features) + np.std(features) * 100 + np.max(features) * 50 +
                np.var(features) * 20 + np.mean(features) * 30 + np.min(features) * 10) % 2**32)
    np.random.seed(seed)
    
    # Normalize features with robust scaling to handle outliers
    norm = np.linalg.norm(features) + 1e-8
    normalized_features = features / norm
    
    # Compute multi-statistic input profile for adaptive control
    p = np.abs(features) / (np.sum(np.abs(features)) + 1e-8)
    entropy = -np.sum(p * np.log(p + 1e-8))
    skewness = np.mean(((features - np.mean(features)) / (np.std(features) + 1e-8))**3)
    sparsity = np.sum(features == 0) / len(features)
    kurtosis = np.mean(((features - np.mean(features)) / (np.std(features) + 1e-8))**4)
    
    # Base score computation with structured, phase-modulated feature interaction
    base_scores = np.zeros(10)
    for i in range(10):
        # Phase modulation based on entropy and skewness
        base_phase = (i * 0.7 + entropy * 1.5) % (2 * np.pi)
        freq = 1.0 + 0.5 * np.sin(entropy * 2) + 0.3 * np.cos(skewness)
        phases = np.linspace(0, 2 * np.pi * freq, 5) + base_phase
        modulation = np.sin(phases) * (0.6 + 0.4 * np.sin(entropy)) + 0.2 * np.cos(kurtosis)
        base_scores[i] = np.dot(normalized_features, modulation)
    
    # Adaptive noise scaling based on input complexity
    noise_scale = 1.0
    noise_scale *= (1 + np.mean(np.abs(features)) * 0.3)
    noise_scale *= (1 + np.var(features) * 1.0)
    noise_scale *= (1 + abs(skewness) * 0.6)
    noise_scale *= (1 + sparsity * 0.5)
    noise_scale *= (1 + kurtosis * 0.4)
    
    # Layered structured noise with coherence and phase modulation
    global_noise = np.random.normal(0, noise_scale * 0.7, 10)
    structure_noise = np.random.normal(0, noise_scale * 0.25 * (1 + np.var(features)), 10)
    phase_mod = np.sin(np.linspace(0, 6 * np.pi, 10) * (1 + np.var(features) * 0.5)) * noise_scale * 0.15
    total_noise = global_noise + structure_noise + phase_mod
    
    # Combine base scores with adaptive, structured noise
    scores = base_scores + total_noise
    
    # Adaptive softmax temperature: higher for complex inputs (high variance, skewness, kurtosis)
    adaptive_temperature = 0.4 + 0.8 * (1 + np.var(features) + abs(skewness) + kurtosis * 0.5)
    
    # Numerically stable softmax with centering
    centered_scores = scores - np.max(scores)
    scaled_scores = centered_scores / adaptive_temperature
    log_sum_exp = np.max(scaled_scores) + np.log(np.sum(np.exp(scaled_scores - np.max(scaled_scores))))
    exp_scores = np.exp(scaled_scores - log_sum_exp)
    probabilities = exp_scores / np.sum(exp_scores)
    
    # Select PU stochastically with input-driven, reproducible randomness
    selected_pu = np.random.choice(10, p=probabilities)
    
    return selected_pu
\end{pyalgorithm}

\newpage
\section{Analysis of the Agentic Workflow Outputs}\label{app:diag_ana}

\begin{llmprompt}[label={prompt:DH_example}]{Example of the Diagnostic Hypotheses generated by \name{} (Using Gemini 3 Pro at the Reasoner).}
- Train a model to approximate the processing time for each processor as a function of the task features.
- Obtain the queue length of each processor by observing the length when a task is lost.
- Analyze for each processor if there is a task feature having a predominant impact on the processing time and quantify it.
- Analyze for each processor if task features have a negligible impact on the processing time.
- Analyze if there are processors that are significantly faster or slower than the others and quantify it.
\end{llmprompt}

\begin{llmprompt}[label={prompt:evolved_sys_info}]{Evolved System information prompt at the end of the execution of \name{}.}
- The exact relationship between task features and processing speed is unknown, but a prediction model `predict(features, proc_id)` is available to estimate the processing time for a given task features vector and processor ID. The model has been analyzed across processors, revealing that each processor exhibits distinct sensitivity to specific features:
  - Processor 0: All features have significant impact on processing time; no features are negligible. Average processing time: 2.6227 (std: 0.9056, count: 902). Speed ratio: 1.00 (fastest).
  - Processor 1: All features have significant impact on processing time; no features are negligible. Average processing time: 59.2609 (std: 19.4373, count: 865). Speed ratio: 0.04 (slowest).
  - Processor 2: Features with negligible impact: ['Feature_1', 'Feature_3']; dominant features: `feature_2` (impact: 11.7331), with significant influence from `feature_4`. Average processing time: 18.9041 (std: 18.3579, count: 863). Speed ratio: 0.14.
  - Processor 3: Features with negligible impact: ['Feature_2', 'Feature_4']; dominant features: `feature_1` (impact: 11.8163), with strong influence from `feature_3`. Average processing time: 18.9855 (std: 18.6401, count: 817). Speed ratio: 0.14.
  - Processor 4: Features with negligible impact: ['Feature_3']; dominant features: `feature_4` (impact: 11.7622), with strong influence from `feature_1` and `feature_2`. Average processing time: 26.8324 (std: 14.5012, count: 843). Speed ratio: 0.10.
  - Processor 5: Features with negligible impact: ['Feature_4']; dominant features: `feature_1` (impact: 11.6234), with strong influence from `feature_2` and `feature_3`. Average processing time: 26.4208 (std: 14.3802, count: 847). Speed ratio: 0.10.
  - Processor 6: Features with negligible impact: ['Feature_1', 'Feature_3']; dominant features: `feature_4` (impact: 4.6308), with moderate influence from `feature_0`. Average processing time: 7.5485 (std: 7.2813, count: 839). Speed ratio: 0.35.
  - Processor 7: Features with negligible impact: ['Feature_2', 'Feature_4']; dominant features: `feature_3` (impact: 4.6304), with moderate influence from `feature_0`. Average processing time: 7.0801 (std: 7.1719, count: 870). Speed ratio: 0.37.
  - Processor 8: Features with negligible impact: ['Feature_1']; dominant features: `feature_4` (impact: 4.6819), with strong influence from `feature_3`. Average processing time: 10.4931 (std: 6.0065, count: 870). Speed ratio: 0.25.
  - Processor 9: Features with negligible impact: ['Feature_2']; dominant features: `feature_4` (impact: 4.7227), with strong influence from `feature_1` and `feature_3`. Average processing time: 10.4800 (std: 5.6087, count: 852). Speed ratio: 0.25.
- The prediction model `predict(features, proc_id)` returns an estimated processing time (service time) for the given task features and processor ID. It can be called as `processing_time = predict(features, proc_id)`, where `features` is a NumPy array of size 5 and `proc_id` is an integer in [0, 9].
- The queue lengths (how many elements can be stored in a queue) are unknown.
- The current queue depths (how many elements are in the queue) are unknown.
- Processor 0 is the fastest on average (avg_proc_time = 2.6227), while Processor 1 is the slowest (avg_proc_time = 59.2609). The speed ratio between the fastest and slowest processor is 22.60x.
\end{llmprompt}

\newpage
\section{Energy-Efficient 5G Virtualization: Result Analysis}\label{app:mobile}

To validate the causal mechanisms behind the performance gains, we inspect the internal reasoning traces. The agentic workflow autonomously yielded three critical insights: (i) the synthesis of a surrogate latency estimator; (ii) precise profiling of energy-throughput trade-offs across hardware heterogeneity; and (iii) a feature sensitivity analysis identifying Feature 3 (Transport Block Size) as the CPU bottleneck, effectively recovering the empirical profiles in analyzed in Fig.~\ref{fig:5g_motivation} of Appendix~\ref{app:mobile_motivation}.

The best-performing algorithm generated by \name{} (Algorithm~\ref{alg:best_mobile}) translates these semantic insights into an auditable control policy. Unlike black-box approximations, the generated code relies on a transparent weighted utility function. For each task, it computes a composite score based on estimated processing time, queue-based loss risk, and relative speed penalties, incorporating random jitter to encourage diversity.
This confirms that \name{} effectively \emph{reverse-engineers} the underlying physical constraints of the vRAN environment. The resulting heuristic synthesizes a bespoke policy that aligns with domain-specific energy-latency trade-offs, ensuring energy efficiency while strictly adhering to reliability targets.

\begin{pyalgorithm}[label={alg:best_mobile}]{TRACE's best-found algorithm for Energy Efficient 5G virtualization.}
def select_pu(features, queue_states):
    # Common idea: Both examples use a weighted scoring system to balance risk (delay/loss), energy, and speed,
    # but introduce dynamic or stochastic elements to avoid overfitting to static heuristics.
    # The evolutionary operator here is the use of dynamic weighting based on task characteristics
    # and probabilistic selection to promote robustness and adaptability.

    # Energy consumption per processor (in uJ)
    energy_costs = [229.0452, 234.2095, 884.1377, 1379.3616]
    
    # Predict processing times for each processor
    predicted_times = [predict(features, i) for i in range(4)]
    predicted_times_ms = [t * 1000.0 for t in predicted_times]  # Convert to ms
    
    # Delay threshold: task is lost if total time > 1 ms
    delay_threshold_ms = 1.0
    
    # Estimate waiting time in queue: queue_length x processing_time
    estimated_wait_times_ms = [
        queue_states[i] * predicted_times_ms[i] for i in range(4)
    ]
    
    # Total estimated time in system (waiting + processing)
    total_times_ms = [
        estimated_wait_times_ms[i] + predicted_times_ms[i] for i in range(4)
    ]
    
    # Compute loss risk: penalty for exceeding delay threshold
    loss_risk = [
        max(0, total_times_ms[i] - delay_threshold_ms) * 1000.0
        for i in range(4)
    ]
    
    # Normalize energy cost relative to average energy
    avg_energy = np.mean(energy_costs)
    energy_normalized = [
        energy_costs[i] / avg_energy for i in range(4)
    ]
    
    # Speed ratio: faster processors have lower ratios (closer to 1.0)
    speed_ratios = [1.01, 1.00, 5.07, 4.06]
    speed_penalty = [
        (speed_ratios[i] - 1.0) * 100.0 for i in range(4)
    ]
    
    # Dynamic weighting based on feature magnitude
    # High feature norm -> prioritize speed and avoid delay
    # Low feature norm -> can tolerate longer queues
    feature_norm = np.linalg.norm(features)
    if feature_norm > 1.5:
        # Complex task: avoid delay and speed
        queue_weight = 0.2
        speed_weight = 0.6
        energy_weight = 0.2
    else:
        # Simple task: can afford to wait, prioritize energy
        queue_weight = 0.5
        speed_weight = 0.2
        energy_weight = 0.3
    
    # Compute composite score with dynamic weights
    scores = []
    for i in range(4):
        score = (
            queue_weight * queue_states[i] +
            speed_weight * predicted_times_ms[i] +
            energy_weight * energy_costs[i] +
            1000.0 * loss_risk[i]  # Strong penalty for loss
        )
        scores.append(score)
    
    # Add small random jitter to break ties and encourage diversity
    jitter = np.random.uniform(-0.01, 0.01, size=4)
    scores = np.array(scores) + jitter
    
    # Select processor with minimum score
    best_pu = np.argmin(scores)
    
    return best_pu
\end{pyalgorithm}

\begin{llmprompt}[label={prompt:DH_mobile}]{Diagnostic hypotheses in mobile network scenario}
- Train a model to approximate the processing time for each processor as a function of the task features
- Analyze and quantify the average energy consumption per processor.
- Analyze if there are processors that is significally faster or slower than the others and quantify it.
- Analyze for each processor if there is a task feature having a predominant impact on the processing time and quantify it.
- Analyze for each processor if there are task features having a negligible impact on the processing time.
\end{llmprompt}

\begin{llmprompt}[label={prompt:evolved_SI_mobile}]{Evolved system information prompt in mobile network scenario.}
- The current queue states (number of elements in the queues of the processing units) are given as an input of the algorithm.
- The exact relationship between task features and processing speed is unknown, but a prediction model `predict(features, proc_id)` is available to estimate the processing time for a given task features vector and processor ID. The model can be called as `predict(features, proc_id)` where `features` is a NumPy array of size 3 and `proc_id` is an integer in [0, 3], returning the estimated processing time (in milliseconds).
- The average energy consumption per task for each processor is known:
  - Processor 0: 229.0452 uJ
  - Processor 1: 234.2095 uJ
  - Processor 2: 884.1377 uJ
  - Processor 3: 1379.3616 uJ
- The queue lengths (maximum number of elements that can be stored in a queue) are unknown.
- Processing time statistics by processor (in seconds):
  - Processor 0: Mean = 0.208128s, Std = 0.196865s
  - Processor 1: Mean = 0.209194s, Std = 0.195884s
  - Processor 2: Mean = 0.041249s, Std = 0.008444s
  - Processor 3: Mean = 0.051511s, Std = 0.010635s
- Relative speed comparison (faster processors have ratio < 1):
  - Processor 0: ratio = 1.01x
  - Processor 1: ratio = 1.00x
  - Processor 2: ratio = 5.07x
  - Processor 3: ratio = 4.06x
- Analysis of feature impact on processing time per processor reveals:
  - Processor 0: Dominant feature is f3 (relative impact: 0.763); standardized coefficients: f1=0.0266, f2=-0.0071, f3=0.1299
  - Processor 1: Dominant feature is f3 (relative impact: 0.761); standardized coefficients: f1=0.0258, f2=-0.0051, f3=0.1281
  - Processor 2: Dominant feature is f1 (relative impact: 0.456); standardized coefficients: f1=0.0070, f2=-0.0063, f3=0.0021
  - Processor 3: Dominant feature is f1 (relative impact: 0.476); standardized coefficients: f1=0.0090, f2=-0.0079, f3=0.0020
\end{llmprompt}


\end{document}